\documentclass[lettersize,journal]{IEEEtran}
\usepackage{amsmath,amsfonts}
\usepackage{algorithmic}
\usepackage{algorithm}
\usepackage{array}
\usepackage[caption=false,font=normalsize,labelfont=sf,textfont=sf]{subfig}
\usepackage{textcomp}
\usepackage{stfloats}
\usepackage{url}
\usepackage{verbatim}
\usepackage{graphicx}
\usepackage{booktabs}
\usepackage{multirow}
\usepackage{tabularx}
\usepackage[table]{xcolor}
\usepackage{cite}
\definecolor{merprow}{RGB}{246,242,251}
\definecolor{merpcodelink}{RGB}{145,105,190}
\begin{document}

\title{MERP: A Latent-Bottleneck Framework for Robust Multi-Modal Multi-Label Emotion Recognition}

\author{Chuangxin Zhao, Meimei Zhang, Canran Xiao, Jiachen Luo, Jiajun He, Kaijie Chen, Huy Phan, Tomoki Toda, and Bj\"orn W. Schuller,~\IEEEmembership{Fellow,~IEEE}%
\thanks{Chuangxin Zhao and Meimei Zhang contributed equally to this work.}%
\thanks{Chuangxin Zhao is with The Hong Kong University of Science and Technology (Guangzhou), Guangzhou, China (e-mail: zhaochuangxin2023@ia.ac.cn).}%
\thanks{Meimei Zhang is with Trine University, Angola, IN, USA (e-mail: mzhang241@my.trine.edu).}%
\thanks{Canran Xiao is with Sun Yat-sen University, Shenzhen Campus, Shenzhen, China (e-mail: xiaocr3@mail.sysu.edu.cn).}%
\thanks{Jiachen Luo is with School of Computation, Information and Technology, Technical University of Munich, Germany, and also with School of Electronical Engineering and Computer Science, Queen Mary University of London, England (e-mail: jiachen.luo@qmul.ac.uk).}%
\thanks{Jiajun He and Tomoki Toda are with Alibaba Group, China, and also with Graduate School of Informatics, Nagoya University, Nagoya, Japan (e-mail: kakashun.hjj@alibaba-inc.com and tomoki@icts.nagoya-u.ac.jp).}%
\thanks{Kaijie Chen is with University of Virginia, USA (e-mail: jkkjjj715@gmail.com).}%
\thanks{Huy Phan is with German Research Center for Artificial Intelligence (DFKI), L\"ubeck, Germany (e-mail: huy.phan@dfki.de).}%
\thanks{Bj\"orn W. Schuller is with CHI -- Chair of Health Informatics, Technical University of Munich, Germany, and also with GLAM -- Group on Language, Audio \& Music, Imperial College London, U.K. (e-mail: schuller@tum.de).}}

\markboth{Zhao \MakeLowercase{\textit{et al.}}}{MERP: Latent-Bottleneck MMER}

\maketitle

\begin{abstract}
Multimodal multi-label emotion recognition is important for affect-aware systems, but models trained with complete text, visual, and acoustic streams often degrade sharply when some streams are unavailable at deployment. We address this train--test gap by learning a compact representation that preserves emotion-specific evidence from any available subset of modalities. We propose the Multi-modal Emotion Recognition Perceiver (MERP), which maps the joint sequence into a fixed-size latent bank, thereby keeping repeated reasoning independent of raw input length; it extracts label-specific evidence before and after latent reasoning and aligns the two levels through reconstruction. MERP further uses positive and negative class prototypes, a noise-gated feature enhancer with frequency-shaped perturbations during training, and confidence-aware class-wise fusion to strengthen separability, reduce sensitivity to missing inputs, and combine complementary predictors. Under 20\% test-time random modality drop, MERP achieves micro-F1 scores of 0.563 on CMU-MOSEI and 0.780 on M\textsuperscript{3}ED, exceeding the strongest evaluated baselines by 0.004 and 0.005, respectively; in the severe CMU-MOSEI audio-only setting, it retains 0.473 micro-F1 versus 0.022 for LDDU. These results show that fixed latent reasoning provides a practical, testable route to robust multimodal emotion recognition without requiring missing-modality examples during training. Code is available at \textcolor{merpcodelink}{\url{https://github.com/Creative-zcx/MERP}}.
\end{abstract}


\section{Introduction}
\IEEEPARstart{U}{nderstanding} human affect from language, voice, and facial behaviour is central to affect-aware interfaces, conversational systems, and large-scale media analysis~\cite{ref1}. These modalities provide complementary evidence, making their joint analysis more informative than isolated unimodal prediction~\cite{ref2}. Multi-modal multi-label emotion recognition (MMER) is especially valuable because an utterance can express several emotions simultaneously, but this setting must resolve both cross-modal ambiguity and correlations among co-occurring labels~\cite{ref3}.

\begin{figure}[!t]
\centering
\includegraphics[width=\columnwidth]{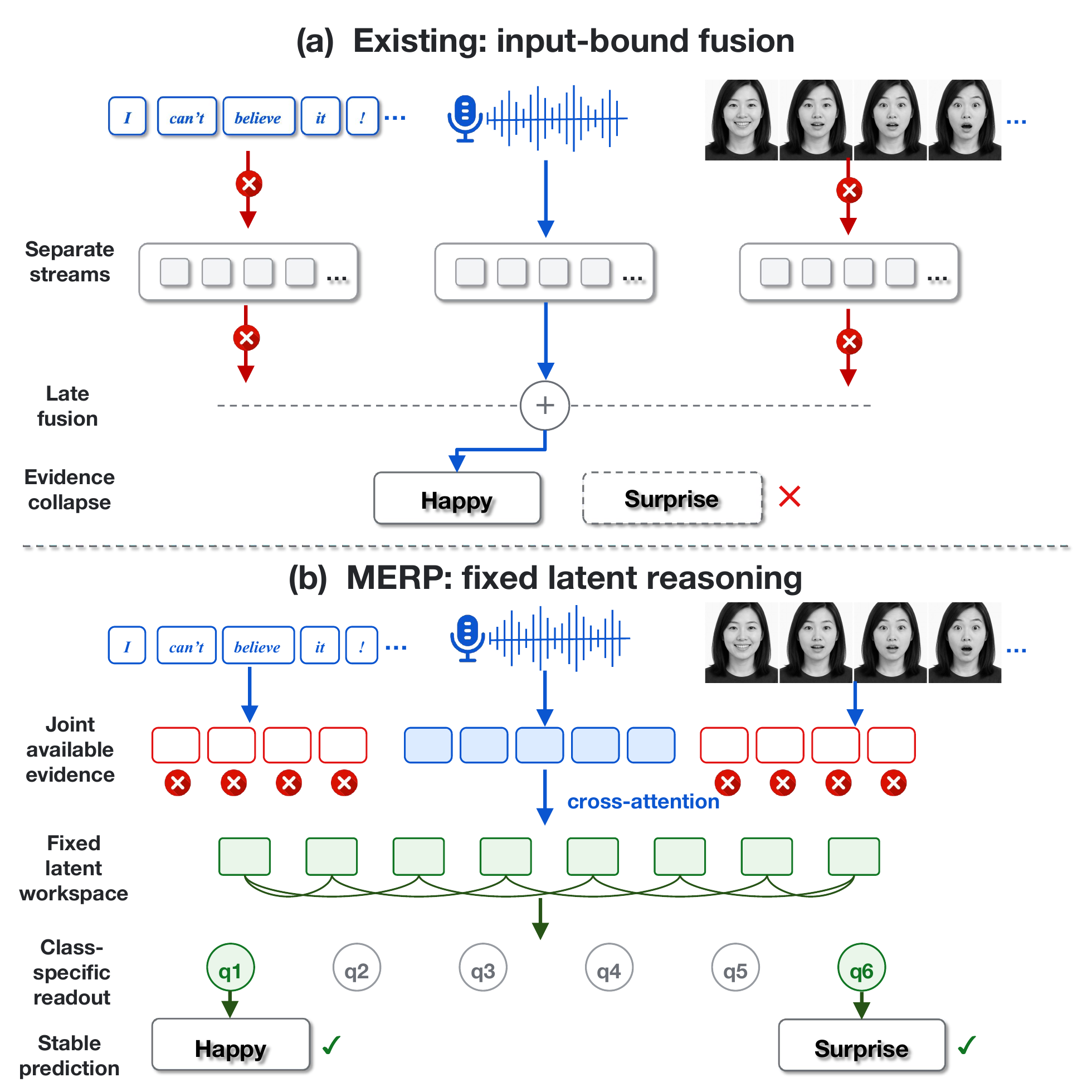}
\caption{\textbf{Fixed latent reasoning preserves class-specific evidence under compound modality loss.} With text and visual streams unavailable, the schematic input-bound path loses the \emph{Surprise} label after late fusion (a). MERP maps the same zeroed input into a fixed latent workspace and uses class-specific readouts to retain both labels (b).}
\label{fig:motivation}
\end{figure}

MMER has advanced substantially through increasingly structured dependency and representation learning. HHMPN propagates information over a hierarchy of modality features and emotion labels~\cite{ref4}, while TAILOR separates common and private modality factors before label-guided decoding~\cite{ref6}. More recent systems improve label discrimination through contrastive reconstruction in CARAT~\cite{ref19} and probabilistic uncertainty modeling in LDDU~\cite{ref13}. In parallel, recent studies have strengthened categorical multi-modal emotion recognition through emotion-shift-aware cross-modal fusion~\cite{ref35}, relational subgraph interaction~\cite{ref36}, and interpretable hybrid fusion~\cite{ref37}. These advances demonstrate the value of structured fusion, but the categorical models assign one emotion to each instance, whereas MMER must distinguish several correlated labels. Moreover, their computation remains organized around modality-specific or joint input sequences. Directional cross-modal attention, for example, captures fine-grained interactions but retains token-level costs that grow with the input streams~\cite{ref5}; label-specific decoding and late probabilistic fusion improve prediction without removing this structural dependence.

Incomplete input exposes a second, closely coupled limitation. Existing approaches recover missing representations through cross-modal imagination~\cite{ref21}, align complete and incomplete views in a shared space~\cite{ref24}, construct a stable proxy modality~\cite{ref30}, or adapt a pretrained model to missing inputs with a small parameter overhead~\cite{ref31}. Recent work further suppresses missing-input noise through heterogeneous graph denoising for conversational emotion recognition~\cite{ref38}. These strategies have broadened the scope of robust multi-modal learning, but they predominantly address single-label sentiment or emotion tasks and usually treat missingness through imputation, alignment, denoising, or missing-pattern adaptation~\cite{ref34}. They therefore do not resolve the central tension in deployable MMER: the model must preserve distinct evidence for multiple correlated emotions as modalities disappear, while its computational burden should not grow with repeated processing of the full input streams. Robustness, label-specific reasoning, and efficiency remain largely separate design objectives.

This work asks whether MMER can retain class-specific evidence under incomplete input within a compact representation whose deep computation is decoupled from the raw sequence length. We propose \textbf{MERP} (Multi-modal Emotion Recognition Perceiver), which follows the latent-query principle of Perceiver IO~\cite{ref9}: available multi-modal evidence is compressed into a fixed latent workspace before iterative reasoning, and emotion-specific evidence is extracted and reinforced across representation levels. This design supports a single predictor for complete and incomplete inputs while bounding the cost of the deeper computation. As illustrated in Figure~\ref{fig:motivation}, MERP replaces input-bound late fusion with fixed-latent, class-specific reasoning under compound text--visual loss.

The main contributions are:
\begin{itemize}
\item We show that robust MMER should not be framed as recovering an absent modality or assigning a single reliability score to each stream. Our depth-wise diagnostic reveals that repeated full-sequence fusion amplifies missingness and corrupts label evidence unevenly across classes and feature channels; the central design objective is therefore to aggregate available cues into a compact workspace and preserve class-specific evidence throughout deeper reasoning.
\item We design MERP around a fixed latent bottleneck, cross-level class-specific readouts, prototype-guided noise-gated enhancement, and confidence-aware class-wise fusion. The same predictor handles complete and incomplete inputs while decoupling deeper reasoning from the raw sequence length.
\item MERP achieves the best micro-F1 under 20\% test-time modality drop on CMU-MOSEI (0.563) and M\textsuperscript{3}ED (0.780), leads LDDU throughout $p_{\text{miss}}\!\in\![0,0.5]$, and avoids its compound-failure collapse, with 5.28\,M parameters and 5.28\,ms latency on CMU-MOSEI.
\end{itemize}

\section{Related Work}\label{sec:related}

\subsection{Multi-Modal Multi-Label Emotion Recognition}
Multi-modal multi-label emotion recognition extends categorical emotion recognition by allowing several affective states to coexist in one utterance. Early work explicitly modeled modality-to-label and label-to-label dependencies to exploit this structure~\cite{ref3}. Transformer-based label-set generation subsequently captured label correlations through sequential decoding~\cite{ref18}, while HHMPN represented modalities and emotions as a heterogeneous hierarchy and propagated information between them~\cite{ref4}. These approaches established the importance of structured label reasoning, although sequential generation imposes a decoding order and hierarchical message passing repeatedly operates on modality-specific features. TAILOR later separated modality-common and modality-private factors before label-guided decoding~\cite{ref6}, and AMP used adversarial masking and perturbation to improve the stability of multi-modal representations~\cite{ref23}.

Recent MMER methods place greater emphasis on discriminative label evidence. CARAT combines contrastive learning with feature reconstruction and aggregation~\cite{ref19}; UniVA transfers valence--arousal structure learned from individual modalities~\cite{ref22}; LDDU represents prediction uncertainty through decoupled latent distributions~\cite{ref13}; and HGE models heterogeneous interactions among modalities and labels~\cite{ref28}. Related advances in categorical conversational emotion recognition capture emotion shifts~\cite{ref35} and inter- and intra-speaker dependencies~\cite{ref36}, but produce one emotion category per instance. Collectively, these methods improve cross-modal interaction and inter-label discrimination, yet their deep computation remains tied to modality streams, and most do not preserve class-specific evidence under single- and multiple-stream loss. MERP addresses both limitations by conducting repeated reasoning in a fixed latent bank and extracting emotion-specific evidence at successive latent depths.

\subsection{Robust Learning with Incomplete Modalities}
Robust multi-modal learning under missing inputs has developed along two complementary directions~\cite{ref34}. Recovery-based methods estimate the information carried by an absent stream. SMIL combines Bayesian meta-learning with feature reconstruction for severe missingness~\cite{ref20}; MMIN imagines missing modality representations from the observed streams~\cite{ref21}; and TFR-Net reconstructs missing temporal segments while learning task representations~\cite{ref26}. Recent main-conference approaches use duplex diffusion to reduce modality-dependent recovery bias~\cite{ref41} or structured knowledge extracted by large multi-modal models for training-free, out-of-distribution completion~\cite{ref42}. Such recovery can restore a complete input interface, but the estimated features may propagate cross-modal prediction error to the downstream classifier. It also introduces auxiliary generation and reconstruction paths whose cost and reliability depend on the missing pattern.

Representation-based methods instead encourage observed subsets to remain predictive. MissModal aligns complete and incomplete views through geometric and distributional constraints~\cite{ref24}, whereas parameter-efficient adaptation modifies a pretrained multi-modal model with few additional parameters~\cite{ref31}. P-RMF constructs a stable proxy modality to reduce the discrepancy between complete and incomplete samples~\cite{ref30}, and contrastive masked autoencoding learns modality-agnostic emotion representations~\cite{ref32}. SimMLM dynamically gates modality experts and regularizes predictions so that additional available modalities do not reduce task performance~\cite{ref40}. More recently, heterogeneous graph denoising has been used to suppress noisy propagation in incomplete multi-modal conversations~\cite{ref38}. These methods substantially improve missing-modality robustness, but they predominantly optimize a global representation for single-label sentiment and emotion prediction; they do not explicitly protect the distinct evidence required by multiple correlated labels. Across this family, training commonly involves simulated missing views, auxiliary completion, and missing-pattern adaptation. MERP uses one predictor for complete and incomplete inputs, avoids synthesizing an absent stream, and maintains class-specific evidence within a shared fixed-size latent space.

\subsection{Latent-Space Multi-Modal Fusion}
Multi-modal fusion architectures differ chiefly in where cross-modal interaction occurs. MulT applies directional cross-attention between unaligned modality sequences, enabling fine-grained temporal exchange at a cost that grows with pairwise token interactions~\cite{ref5}. Common--private factorization in MISA compresses each modality into invariant and modality-specific components before fusion~\cite{ref17}, while VISTANet combines intermediate and late fusion with adaptive modality weighting for interpretable categorical prediction~\cite{ref37}. These designs retain useful modality structure, but their fusion computation remains coupled to the number and length of the input streams.

Latent-bottleneck architectures decouple repeated processing from raw input length. Perceiver IO first cross-attends from a fixed latent array to arbitrary inputs and then performs deep self-attention only among the latent tokens~\cite{ref9}. The Multimodal Bottleneck Transformer similarly restricts audio--visual exchange to a small collection of fusion tokens~\cite{ref27}. Recent robust-fusion methods further model unimodal aleatoric uncertainty~\cite{ref29} and learn an information-bottleneck latent space shared by complete and incomplete inputs~\cite{ref33}. Nevertheless, these architectures target general perception and single-label sentiment tasks and provide no mechanism ensuring that the compact space retains separable evidence for every co-occurring emotion. MERP connects fixed-size latent computation to label-attentive readouts, cross-level consistency, and class-wise fusion, thereby making the bottleneck directly support robust MMER rather than only global multi-modal compression.

\section{Preliminary Study}\label{sec:preliminary}

Before introducing MERP, we examine why a multi-modal fusion model trained with complete observations becomes fragile under modality loss. We conduct a controlled diagnostic study with a conventional dense-fusion Transformer, isolating failure patterns of the fusion process from the components introduced in our framework.

\subsection{Diagnostic Protocol}
\noindent\textbf{Probe model.}
We train a six-layer Transformer fusion model on the CMU-MOSEI training split using only complete multi-modal inputs. Text, visual, and acoustic features are independently projected to a common hidden dimension and concatenated into one sequence, followed by six self-attention layers and a multi-label classification head. The probe excludes latent queries, cross-level reconstruction, class prototypes, feature-noise enhancement, and adaptive branch fusion.

All reported diagnostic metrics are computed on the CMU-MOSEI validation split with the probe parameters frozen; the training split is used only to fit the probe and estimate the clean class centroids. For each validation utterance $\mathbf{x}$, we evaluate the intact input and six corrupted views,
\begin{equation}
\mathcal{S}=\{\text{drop-T},\text{drop-V},\text{drop-A},
\text{T-only},\text{V-only},\text{A-only}\},
\end{equation}
where drop-$m$ removes modality $m$, and $m$-only retains only modality $m$. Removed streams are replaced by zeros, matching the modality-drop operation in Eq.~\eqref{eq:missing-input}.

\noindent\textbf{Depth-wise representation retention.}
Let $\mathbf{H}^{(\ell)}$ and $\widetilde{\mathbf{H}}_{s}^{(\ell)}$ denote the sample representations of the intact and corrupted validation sets after fusion layer $\ell$ under corruption pattern $s$. Their agreement is measured using linear centered-kernel alignment (CKA)~\cite{ref39}:
\begin{equation}
R_{\ell,s}=\operatorname{CKA}\!\left(
\mathbf{H}^{(\ell)},\widetilde{\mathbf{H}}_{s}^{(\ell)}
\right).
\label{eq:prelim-cka}
\end{equation}
A smaller $R_{\ell,s}$ indicates a larger corruption-induced change in the representation.

We also measure the attention assigned to unavailable tokens. Let $\mathcal{I}_{s}$ contain the positions of the removed modalities, and let $\widetilde{\mathbf{A}}_{s}^{(\ell)}$ be the corrupted-view attention matrix averaged over heads and validation samples. The unavailable-token attention ratio is
\begin{equation}
U_{\ell,s}=
\frac{N_{\mathrm{q}}^{-1}\sum_{i=1}^{N_{\mathrm{q}}}
\sum_{j\in\mathcal{I}_{s}}\widetilde{A}_{s,ij}^{(\ell)}}
{|\mathcal{I}_{s}|/T_{\Sigma}}.
\label{eq:prelim-uar}
\end{equation}
The denominator is the attention share under a uniform distribution. Thus, $U_{\ell,s}=1$ corresponds to uniform attention, and values above one indicate disproportionate attention to positions with no modality content.

\noindent\textbf{Class-wise geometry and prediction stability.}
Using clean training representations from the frozen probe, we compute positive and negative centroids $\boldsymbol{\mu}_{c}^{+}$ and $\boldsymbol{\mu}_{c}^{-}$ for each emotion $c$, with $\boldsymbol{\mu}_{c}^{1}=\boldsymbol{\mu}_{c}^{+}$ and $\boldsymbol{\mu}_{c}^{0}=\boldsymbol{\mu}_{c}^{-}$. For a representation $\mathbf{h}$ with binary label $y_c$, the class-separation margin is
\begin{equation}
m_c(\mathbf{h},y_c)=
\cos(\mathbf{h},\boldsymbol{\mu}_{c}^{y_c})-
\cos(\mathbf{h},\boldsymbol{\mu}_{c}^{1-y_c}).
\end{equation}
The relative margin collapse under corruption $s$ is
\begin{equation}
\Delta M_{c,s}=100\!\left(
1-\frac{\mathbb{E}[m_c(\widetilde{\mathbf{h}}_s,y_c)]}
{\mathbb{E}[m_c(\mathbf{h},y_c)]}
\right).
\label{eq:prelim-margin}
\end{equation}
Prediction instability is quantified by the clean-to-corrupted label flip rate
\begin{equation}
F_{c,s}=\frac{100}{N}\sum_{b=1}^{N}
\mathbb{I}\!\left[\widehat{y}_{b,c}\neq
\widetilde{y}_{b,c,s}\right].
\label{eq:prelim-flip}
\end{equation}
Finally, we form the class-conditioned perturbation
\begin{equation}
\boldsymbol{\delta}_{b,c,s}=
(\widetilde{\mathbf{h}}_{b,s}-\mathbf{h}_{b})\odot\mathbf{w}_{c},
\end{equation}
where $\mathbf{w}_{c}$ is the class-$c$ classifier weight, and report the fraction of its energy contained in the largest 10\% of feature channels.

\begin{figure}[!t]
\centering
\includegraphics[width=\columnwidth]{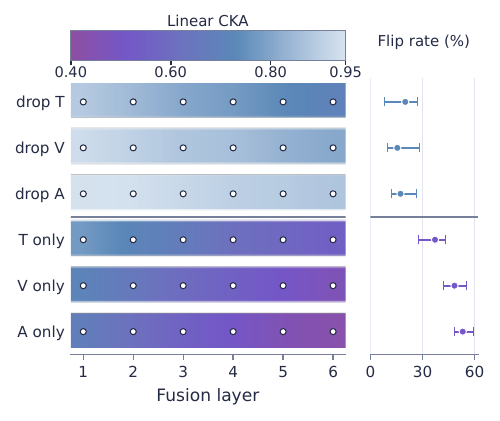}
\caption{Depth-wise missing-modality diagnosis on the CMU-MOSEI validation split. Each colored strip reports linear CKA across all six fusion layers for one corruption pattern, with white-edged dots marking the measured layer outputs. Color is interpolated only along depth within each categorical row. The aligned marginal plot reports the mean and range of class-wise flip rates. The horizontal divider separates single-modality drops from single-retained inputs.}
\label{fig:prelim-diagnosis}
\end{figure}

\subsection{Findings and Design Implications}
\noindent\textbf{Modality loss is amplified with fusion depth.}
Figure~\ref{fig:prelim-diagnosis} shows a progressive shift toward lower CKA across all six fusion layers and every corruption pattern. Averaged over the three single-modality drops, CKA decreases from 0.921 at layer 1 to 0.794 at layer 6. When only one modality remains, the decrease is larger, from 0.720 to 0.480. Meanwhile, the unavailable-token attention ratio increases from 1.049 to 1.240 for single-modality drops and from 1.171 to 1.595 for single-retained views. Repeated full-sequence fusion therefore amplifies a local input failure by continually mixing the surviving evidence with positions whose modality content has been removed.

\noindent\textbf{The failure is class-structured and feature-concentrated.}
The layer-6 CKA values are strongly anticorrelated with the mean label flip rate across the six corruption patterns ($r=-0.96$). The aligned ranges in Fig.~\ref{fig:prelim-diagnosis} further show that emotions subjected to the same corruption differ by 10.9--19.4 percentage points in flip rate. Across all 36 emotion--corruption pairs, margin collapse and label instability are strongly associated ($r=0.94$). The most affected 10\% of classifier-relevant channels contain 46.1--87.8\% of the perturbation energy, with a mean of 65.5\%. A global modality reliability score therefore cannot represent either the class variation or the concentration of corruption within the feature space.

These findings impose three requirements on robust MMER. Repeated reasoning must operate on a compact representation after the available evidence has been aggregated, limiting depth-wise error amplification. Emotion-specific evidence must remain identifiable as the representation evolves. Feature enhancement and prediction fusion must respond to class- and feature-dependent corruption. MERP instantiates these requirements through a fixed latent workspace, cross-level label readouts, prototype-guided feature organization, selective enhancement, and class-wise fusion.

\section{The MERP Framework}\label{sec:method}

\subsection{Problem Setup and Notation}
Let $\mathcal{M}=\{t,v,a\}$ denote the text, visual, and acoustic modalities. A mini-batch contains feature sequences $\mathbf{X}_m\in\mathbb{R}^{B\times T_m\times d_m}$ for $m\in\mathcal{M}$ and binary emotion labels $\mathbf{Y}\in\{0,1\}^{B\times C}$, where $B$ is the batch size, $T_m$ is the modality-specific sequence length, $d_m$ is the input dimension, and $C$ is the number of emotion classes. The experiments use $T_m{=}60$ and $C{=}6$. An availability indicator $a_{b,m}\in\{0,1\}$ specifies whether modality $m$ is observed for sample $b$. Missing content is represented by
\begin{equation}
\bar{\mathbf{X}}_{b,m}=a_{b,m}\mathbf{X}_{b,m}.
\label{eq:missing-input}
\end{equation}
All indicators equal one during training. At test time, the indicators are set by the random-drop and fixed-combination protocols in Sections~\ref{sec:experiments} and~\ref{sec:per-modality}. The task is to estimate class probabilities $\mathbf{P}\in[0,1]^{B\times C}$ from the available multi-modal evidence.

\subsection{Architecture Overview}
\begin{figure*}[!t]
\centering
\includegraphics[width=\textwidth]{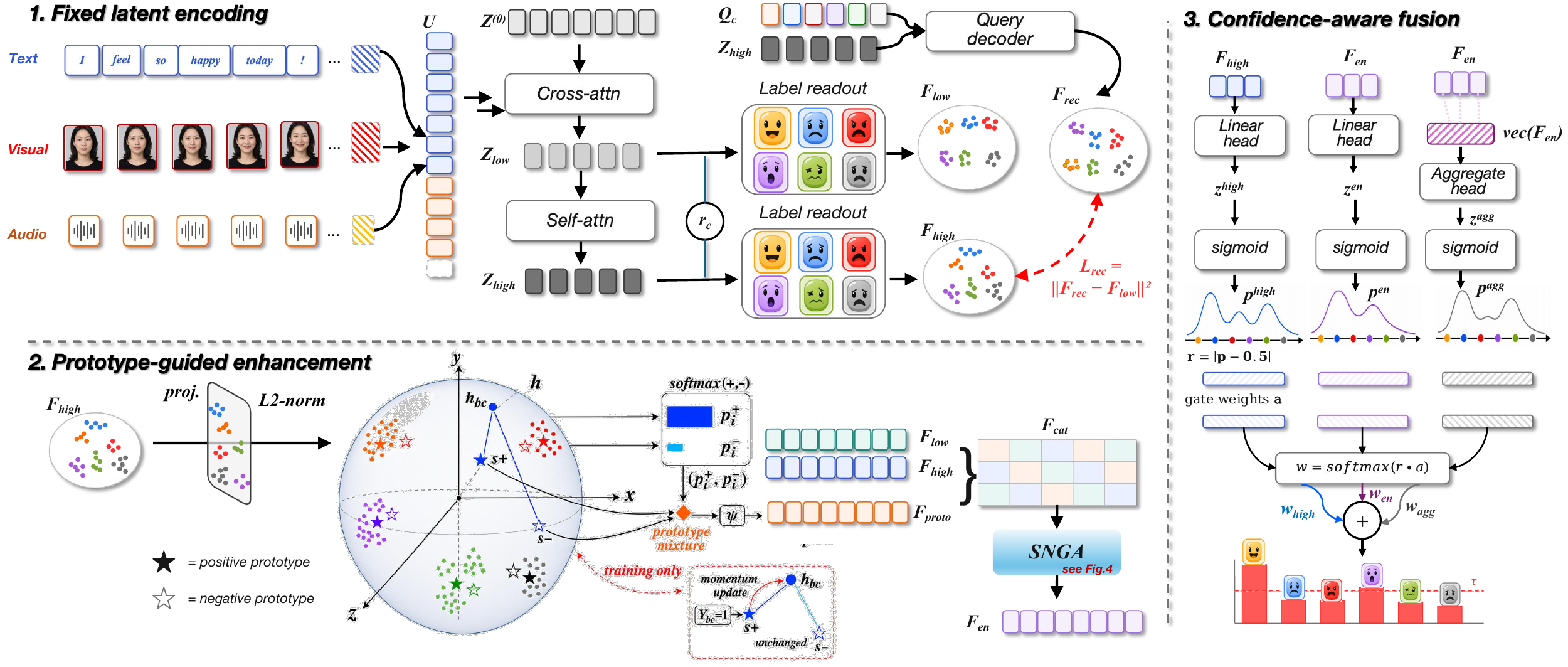}
\caption{Overall computation pipeline of MERP. (1) Joint multimodal tokens $\mathbf{U}$ are compressed by learnable latent queries $\mathbf{Z}^{(0)}$ into $\mathbf{Z}_{\mathrm{low}}$ and refined to $\mathbf{Z}_{\mathrm{high}}$. Class-specific readouts yield $\mathbf{F}_{\mathrm{low}}$ and $\mathbf{F}_{\mathrm{high}}$, while a class-query decoder reconstructs $\mathbf{F}_{\mathrm{rec}}$ and enforces cross-level consistency. (2) Normalised high-level features are matched to positive and negative class prototypes to form $\mathbf{F}_{\mathrm{proto}}$; the prototype, low-level, and high-level features are concatenated and enhanced by SNGA to obtain $\mathbf{F}_{\mathrm{en}}$. (3) High-level, enhanced, and aggregate probabilities are fused class-wise using learned gate preferences calibrated by $r=|p-0.5|$, then thresholded for multi-label prediction.}
\label{fig:merp-arch}
\end{figure*}

\begin{figure}[!t]
\centering
\includegraphics[width=\columnwidth]{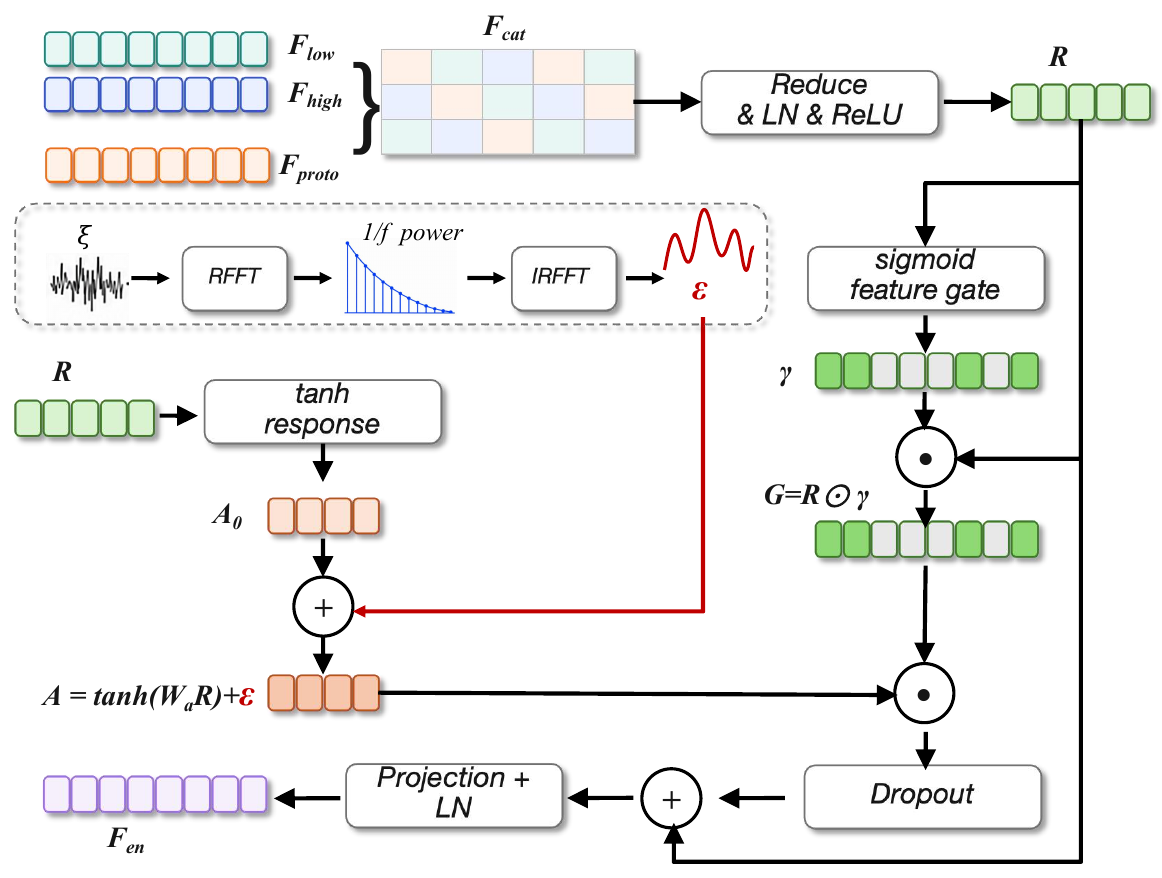}
\caption{Architecture of the spectral noise-gated enhancement (SNGA) block. The concatenated class features $\mathbf{F}_{\mathrm{cat}}$ are reduced to $\mathbf{R}$. The response branch filters white noise $\boldsymbol{\xi}$ in the Fourier domain to obtain a training-only $1/f$ perturbation $\boldsymbol{\epsilon}$, whereas the gating branch computes $\boldsymbol{\gamma}$ and $\mathbf{G}=\mathbf{R}\odot\boldsymbol{\gamma}$. The modulated response is regularised by dropout, added residually to $\mathbf{R}$, and projected to $\mathbf{F}_{\mathrm{en}}$; at inference, $\boldsymbol{\epsilon}=\mathbf{0}$.}
\label{fig:snga-detail}
\end{figure}

Figure~\ref{fig:merp-arch} presents the complete computation flow. After aligning and gating the available modalities, MERP maps the joint sequence to low- and high-level fixed-size latent banks. Class-specific readouts produce $\mathbf{F}_{\mathrm{low}}$ and $\mathbf{F}_{\mathrm{high}}$, while a query decoder reconstructs the low-level evidence from $\mathbf{Z}_{\mathrm{high}}$. The high-level features are then shaped by positive and negative class prototypes, concatenated with the low-level and prototype-conditioned features, and enhanced by SNGA. Finally, the high-level, enhanced, and aggregate probability branches are combined by confidence-aware fusion. Figure~\ref{fig:snga-detail} expands only the internal SNGA computation; complete layer definitions and auxiliary objectives are given in Appendices~\ref{app:preprocessing}--\ref{app:objective}.
Algorithm~\ref{alg:merp} summarises the corresponding end-to-end training and inference procedure.

\subsection{Multi-Modal Preprocessing and Latent Encoding}\label{sec:method-encoder}
For modality $m$, layer-normalised content features are concatenated with a one-dimensional Fourier position code $\mathbf{P}_m$ and padded with trainable channel embeddings to a common width $D$. Denoting the resulting tensor by $\bar{\mathbf{E}}_m\in\mathbb{R}^{B\times T_m\times D}$, the modality gate is
\begin{equation}
\mathbf{R}_m=\mathbf{W}_m\bar{\mathbf{E}}_m+\mathbf{b}_m,\qquad
\mathbf{U}_m=\sigma(\mathbf{R}_m)\odot\mathbf{R}_m.
\label{eq:modality-gate}
\end{equation}
The gate parameters are modality-specific. The joint input is
\begin{equation}
\mathbf{U}=\operatorname{Concat}_{m\in\mathcal{M}}(\mathbf{U}_m)
\in\mathbb{R}^{B\times T_{\Sigma}\times D},\qquad
T_{\Sigma}=\sum_{m\in\mathcal{M}}T_m.
\label{eq:joint-input}
\end{equation}
The exact Fourier construction and channel-alignment operation are specified in Appendix~\ref{app:preprocessing}.

MERP maintains a trainable latent bank $\mathbf{Z}^{(0)}\in\mathbb{R}^{L\times H}$, which is broadcast across the batch. A cross-attention block maps the joint input to a low-level latent representation,
\begin{equation}
\mathbf{Z}_{\mathrm{low}}=\operatorname{CrossAttn}(\mathbf{Z}^{(0)},\mathbf{U})
\in\mathbb{R}^{B\times L\times H}.
\label{eq:latent-cross}
\end{equation}
The final latent representation is obtained with $N=N_{\mathrm{blocks}}N_{\mathrm{self}}$ self-attention blocks,
\begin{equation}
\mathbf{Z}_{\mathrm{high}}=\operatorname{SelfAttn}^{N}(\mathbf{Z}_{\mathrm{low}})
\in\mathbb{R}^{B\times L\times H}.
\label{eq:latent-self}
\end{equation}
We use $L{=}80$, $H{=}256$, $N_{\mathrm{blocks}}{=}1$, and $N_{\mathrm{self}}{=}6$. The attention cost is $O(LT_{\Sigma}+NL^2)$: only the first block depends on the input length, while all deeper reasoning operates on $L$ latent tokens. Under modality drop, Eq.~\eqref{eq:missing-input} removes the designated content stream and the same fixed-size latent computation is applied to the remaining evidence.

\subsection{Label-Specific Readout and Cross-Level Consistency}\label{sec:method-readout}
Let $\mathbf{r}_c\in\mathbb{R}^{H}$ be the trainable readout vector for class $c$. For representation level $q\in\{\mathrm{low},\mathrm{high}\}$, label attention is computed over the $L$ latent tokens:
\begin{equation}
\alpha^{q}_{b,c,i}=
\frac{\exp(\mathbf{r}_c^{\top}\mathbf{Z}_{q,b,i})}
{\sum_{j=1}^{L}\exp(\mathbf{r}_c^{\top}\mathbf{Z}_{q,b,j})},\qquad
\mathbf{F}_{q,b,c}=\sum_{i=1}^{L}\alpha^{q}_{b,c,i}\mathbf{Z}_{q,b,i}.
\label{eq:label-readout}
\end{equation}
Thus $\mathbf{F}_{\mathrm{low}},\mathbf{F}_{\mathrm{high}}\in\mathbb{R}^{B\times C\times H}$. A shared class-specific classifier produces auxiliary logits
\begin{equation}
z^{q}_{b,c}=\mathbf{w}_c^{\top}\mathbf{F}_{q,b,c},\qquad q\in\{\mathrm{low},\mathrm{high}\}.
\label{eq:readout-logits}
\end{equation}

To preserve class-specific evidence across latent depth, MERP uses $C$ trainable decoder queries $\mathbf{Q}\in\mathbb{R}^{C\times H}$. Cross-attention from these queries to $\mathbf{Z}_{\mathrm{high}}$ yields $\mathbf{F}_{\mathrm{rec}}\in\mathbb{R}^{B\times C\times H}$, and the cross-level reconstruction loss is
\begin{equation}
\mathcal{L}_{\mathrm{rec}}=
\frac{1}{BCH}\left\|\mathbf{F}_{\mathrm{rec}}-\mathbf{F}_{\mathrm{low}}\right\|_F^2.
\label{eq:reconstruction}
\end{equation}
This objective aligns decoded high-level semantics with the class-specific evidence extracted immediately after input-to-latent aggregation. Decoder and attention-block details appear in Appendix~\ref{app:attention}.

\subsection{Prototype-Guided Feature Enhancement}\label{sec:method-enhance}
The high-level class features are projected and normalised,
\begin{equation}
\mathbf{h}_{b,c}=\frac{\phi(\mathbf{F}_{\mathrm{high},b,c})}
{\|\phi(\mathbf{F}_{\mathrm{high},b,c})\|_2}
\in\mathbb{R}^{d_z}.
\label{eq:projected-feature}
\end{equation}
Each class maintains a positive prototype $\boldsymbol{\mu}_{c}^{+}$ and a negative prototype $\boldsymbol{\mu}_{c}^{-}$. Their similarity to $\mathbf{h}_{b,c}$ defines a two-way distribution
\begin{equation}
\pi^{s}_{b,c}=
\frac{\exp(\mathbf{h}_{b,c}^{\top}\boldsymbol{\mu}_{c}^{s})}
{\sum_{u\in\{+,-\}}\exp(\mathbf{h}_{b,c}^{\top}\boldsymbol{\mu}_{c}^{u})},\qquad s\in\{+,-\},
\label{eq:prototype-mixture}
\end{equation}
from which the prototype-conditioned feature is
\begin{equation}
\mathbf{F}_{\mathrm{proto},b,c}=
\psi\!\left(\sum_{s\in\{+,-\}}\pi^{s}_{b,c}\boldsymbol{\mu}_{c}^{s}\right)
\in\mathbb{R}^{H}.
\label{eq:prototype-feature}
\end{equation}
The prototypes are updated from labelled training samples, and the projected high-level and enhanced features are regularised with supervised contrastive learning~\cite{ref10} and a memory queue. Appendix~\ref{app:prototypes} gives the update rule, queue construction, and contrastive objective.

The enhancement input concatenates the prototype, low-level, and high-level class features:
\begin{equation}
\mathbf{F}_{\mathrm{cat}}=
[\mathbf{F}_{\mathrm{proto}};\mathbf{F}_{\mathrm{low}};\mathbf{F}_{\mathrm{high}}]
\in\mathbb{R}^{B\times C\times 3H}.
\label{eq:enhance-input}
\end{equation}
A spectral noise-gated feature block maps $\mathbf{F}_{\mathrm{cat}}$ to $\mathbf{F}_{\mathrm{en}}\in\mathbb{R}^{B\times C\times H}$. It injects normalised $1/f$ noise along the feature axis during training, applies a sigmoid feature gate, and projects the residual representation to width $H$. No noise is applied at inference. The complete block is defined in Appendix~\ref{app:snga}. The enhanced class logit is
\begin{equation}
z^{\mathrm{en}}_{b,c}=\mathbf{w}_c^{\top}\mathbf{F}_{\mathrm{en},b,c}.
\label{eq:enhanced-logit}
\end{equation}
An aggregate head additionally maps $\operatorname{vec}(\mathbf{F}_{\mathrm{en}})$ to $\mathbf{z}^{\mathrm{agg}}\in\mathbb{R}^{B\times C}$, retaining dependencies among emotion classes.

\subsection{Class-Wise Fusion and Training Objective}\label{sec:method-objective}
The final prediction combines the aggregate, high-level, and enhanced branches. For $j\in\mathcal{J}=\{\mathrm{agg},\mathrm{high},\mathrm{en}\}$, let $p^j_{b,c}=\sigma(z^j_{b,c})$ and $r^j_{b,c}=|p^j_{b,c}-0.5|$. A class-wise gate receives the three branch probabilities and returns branch preferences
\begin{equation}
\mathbf{a}_{b,c}=\operatorname{softmax}_{j}\!\left(
G_{\theta}([p^{\mathrm{agg}}_{b,c},p^{\mathrm{high}}_{b,c},p^{\mathrm{en}}_{b,c}])
\right).
\label{eq:branch-gate}
\end{equation}
Confidence-calibrated fusion weights and the final probability are
\begin{equation}
\omega^j_{b,c}=\operatorname{softmax}_{j}(r^j_{b,c}a^j_{b,c}),\qquad
p^{\mathrm{final}}_{b,c}=\sum_{j\in\mathcal{J}}\omega^j_{b,c}p^j_{b,c}.
\label{eq:score-fusion}
\end{equation}
The gate is trained through binary cross-entropy on $\mathbf{P}^{\mathrm{final}}$. Predictions are $\hat{Y}_{b,c}=\mathbb{I}[p^{\mathrm{final}}_{b,c}\geq\tau]$, with $\tau{=}0.25$.

The training objective is
\begin{equation}
\begin{aligned}
\mathcal{L}={}&\mathcal{L}_{\mathrm{final}}+\mathcal{L}_{\mathrm{sup}}
+\lambda_{\mathrm{rec}}\mathcal{L}_{\mathrm{rec}}
+\lambda_{\mathrm{cl}}\mathcal{L}_{\mathrm{cl}}\\
&+\lambda_{\mathrm{info}}\mathcal{L}_{\mathrm{info}}
+\lambda_{\mathrm{shuf}}\mathcal{L}_{\mathrm{shuf}}
+\lambda_{\mathrm{reg}}\mathcal{L}_{\mathrm{reg}},
\end{aligned}
\label{eq:total-objective}
\end{equation}
where $\mathcal{L}_{\mathrm{final}}$ supervises the fused probabilities, $\mathcal{L}_{\mathrm{sup}}$ collects logit-level supervision for the low, high, enhanced, and aggregate heads, $\mathcal{L}_{\mathrm{cl}}$ is the queue-based supervised contrastive loss, $\mathcal{L}_{\mathrm{info}}$ aligns $\mathbf{F}_{\mathrm{en}}$ with $\mathbf{F}_{\mathrm{low}}$, $\mathcal{L}_{\mathrm{shuf}}$ regularises class-wise aggregation, and $\mathcal{L}_{\mathrm{reg}}$ penalises the enhancement-block weights. Appendix~\ref{app:objective} provides the complete formulas and coefficients.

\begin{algorithm}[!t]
\caption{End-to-end training and inference of MERP}
\label{alg:merp}
{\footnotesize
\begin{algorithmic}[1]
\REQUIRE Streams $\{\mathbf{X}_m\}_{m\in\mathcal{M}}$, indicators $\{a_{b,m}\}$, parameters $\Theta$, prototype buffers $\boldsymbol{\mu}^{\pm}$, memory queue $\mathcal{Q}$; labels $\mathbf{Y}$ during training
\ENSURE Fused probabilities $\mathbf{P}^{\mathrm{final}}$; predictions $\hat{\mathbf{Y}}$ at inference
\STATE Zero unavailable content; align, gate, and concatenate the streams to form $\mathbf{U}$ using Eqs.~\eqref{eq:missing-input}--\eqref{eq:joint-input}
\STATE $\mathbf{Z}_{\mathrm{low}}\leftarrow\operatorname{CrossAttn}(\mathbf{Z}^{(0)},\mathbf{U})$
\STATE $\mathbf{Z}_{\mathrm{high}}\leftarrow\operatorname{SelfAttn}^{N}(\mathbf{Z}_{\mathrm{low}})$
\FOR{$q\in\{\mathrm{low},\mathrm{high}\}$}
    \STATE $(\mathbf{F}_{q},\mathbf{z}^{q})\leftarrow\operatorname{LabelReadout}(\mathbf{Z}_{q})$
\ENDFOR
\STATE $\mathbf{F}_{\mathrm{rec}}\leftarrow\operatorname{CrossAttn}(\mathbf{Q},\mathbf{Z}_{\mathrm{high}})$
\STATE Project $\mathbf{F}_{\mathrm{high}}$; construct $\mathbf{F}_{\mathrm{proto}}$ from $\boldsymbol{\mu}^{\pm}$ by soft mixing in training and hard selection at inference
\STATE $\mathbf{F}_{\mathrm{en}}\leftarrow\operatorname{SNGA}([\mathbf{F}_{\mathrm{proto}};\mathbf{F}_{\mathrm{low}};\mathbf{F}_{\mathrm{high}}])$, with noise disabled at inference
\STATE Obtain $\mathbf{z}^{\mathrm{en}}$ and $\mathbf{z}^{\mathrm{agg}}$; fuse the three probability branches by Eqs.~\eqref{eq:branch-gate}--\eqref{eq:score-fusion}
\IF{training}
    \STATE Compute $\mathcal{L}_{\mathrm{rec}}$ and the total objective $\mathcal{L}$ in Eq.~\eqref{eq:total-objective}; update $\Theta$ with AdamW
    \STATE Update $\boldsymbol{\mu}^{\pm}$ with labelled high-level features and enqueue the high-level and enhanced views in $\mathcal{Q}$
\ELSE
    \STATE $\hat{Y}_{b,c}\leftarrow\mathbb{I}[p^{\mathrm{final}}_{b,c}\geq\tau]$
\ENDIF
\RETURN $\mathbf{P}^{\mathrm{final}}$ and, at inference, $\hat{\mathbf{Y}}$
\end{algorithmic}
}
\end{algorithm}

\section{Experiments}\label{sec:experiments}
\subsection{Experimental Setup}\label{sec:experimental-setup}
\noindent\textbf{Datasets.}
We evaluate on \textbf{CMU-MOSEI}~\cite{ref11} (16{,}326 / 1{,}871 / 4{,}659 train/valid/test utterances; English; six-class multi-label) and \textbf{M\textsuperscript{3}ED}~\cite{ref12} (Chinese conversational corpus; 685 / 126 / 179 train/valid/test dialogues with utterance-level annotations mapped to the same six classes). Pre-extracted features follow common practice~\cite{ref19}: GloVe (300d) for text, FACET (35d) for visual, and COVAREP (74d) for audio on CMU-MOSEI; RoBERTa (768d), DenseFace (1024d), and Wav2Vec (342d) for M\textsuperscript{3}ED.

\noindent\textbf{Test-time modality-drop protocol.}
We evaluate every method under a unified test-time modality-drop protocol. With probability $p_{\text{miss}}$, one uniformly sampled modality of each selected test example is replaced by zeros; the other two modalities remain intact. The corresponding sequence mask is also zeroed for baseline implementations that consume modality masks. MERP receives the zeroed feature tensor defined in Eq.~\eqref{eq:missing-input}. The decisions are drawn once with a fixed RNG (seed 1) and reused across all methods, so every method sees the same corrupted examples and the same per-example modality choice. Training uses complete inputs, $p_{\text{miss}}{=}0$. Unless stated otherwise, we report $p_{\text{miss}}{=}0.2$ (900 of 4{,}659 CMU-MOSEI test utterances and 43 of 179 M\textsuperscript{3}ED test dialogues are corrupted).

\noindent\textbf{Implementation details.}
MERP uses the Perceiver-IO fusion backbone described in Section~\ref{sec:method-encoder}, with $L{=}80$ latents, $H{=}256$, $N_{\text{blocks}}{=}1$, $N_{\text{self}}{=}6$, one cross-attention head, and eight self-attention heads. We optimise with AdamW for 35 epochs at a learning rate of $3.7{\times}10^{-5}$ with cosine decay (warmup ratio 0.15), batch size 64, dropout 0.18, and binary classification threshold $\tau{=}0.25$. Appendix~\ref{app:objective} reports all objective coefficients. All baselines are retrained from their official code on identical features, splits, and preprocessing. The 2025 state-of-the-art LDDU~\cite{ref13} does not release pre-trained weights; we therefore retrain it under our hardware budget (batch size 16 and queue size 512 on CMU-MOSEI; batch size 8 and queue size 256 on M\textsuperscript{3}ED). GPU memory is measured during single-utterance inference on one NVIDIA A100; per-sample latency is averaged over 200 forward passes after a 20-step warm-up.

\subsection{Main Results under Modality Drop}\label{sec:main}
Table~\ref{tab:main_results} reports all metrics under $p_{\text{miss}}{=}0.2$ on CMU-MOSEI and M\textsuperscript{3}ED. All baselines are retrained from their official releases and evaluated with the same per-sample missing schedule. Parameter counts marked by $^{*}$ for the Transformer-fusion baselines (HHMPN, TAILOR, CARAT, MISA, MulT, and AMP) are taken from prior measurements using their official implementations~\cite{ref19}; GPU memory for all baselines is measured by us under the same inference setup as MERP and LDDU.

\begin{table*}[!t]
\caption{Performance under 20\% test-time random modality drop ($p_{\text{miss}}{=}0.2$, seed=1) on CMU-MOSEI and M\textsuperscript{3}ED. Acc is example-based Jaccard accuracy; P and R are micro-averaged over all sample--label decisions, and $F_{\beta}=(1+\beta^2)PR/(\beta^2P+R)$ (micro-F1 uses $\beta{=}1$). Within each dataset, the best result in each column is in \textbf{bold} and the second-best is \underline{underlined}; higher is better except for Params and GPU Mem. MERP rows are highlighted in very light purple.}
\label{tab:main_results}
\centering
\small
\setlength{\tabcolsep}{2pt}
\renewcommand{\arraystretch}{1.08}
\begin{tabularx}{\textwidth}{@{}l@{\hspace{8pt}}
>{\hsize=1.4\hsize\raggedright\arraybackslash}X
>{\hsize=.7\hsize\centering\arraybackslash}X
>{\hsize=.7\hsize\centering\arraybackslash}X
>{\hsize=.7\hsize\centering\arraybackslash}X
>{\hsize=.8\hsize\centering\arraybackslash}X
>{\hsize=.8\hsize\centering\arraybackslash}X
>{\hsize=1.0\hsize\centering\arraybackslash}X
>{\hsize=1.1\hsize\centering\arraybackslash}X
>{\hsize=1.8\hsize\centering\arraybackslash}X@{}}
\toprule
\textbf{Dataset} & \textbf{Method} & \textbf{Acc} & \textbf{P} & \textbf{R} & $\mathbf{F_{0.5}}$ & $\mathbf{F_2}$ & \textbf{micro-F1} & \textbf{Params} & \textbf{GPU Mem.} \\
\midrule
 & HHMPN~\cite{ref4}    & 0.380 & 0.554 & 0.422 & 0.521 & 0.443 & 0.479 & 32.5\,M$^{*}$ & $>$15\,360\,MiB \\
 & TAILOR~\cite{ref6}   & 0.333 & 0.553 & 0.374 & 0.505 & 0.400 & 0.446 & 14.6\,M$^{*}$ & 6\,518\,MiB \\
 & CARAT~\cite{ref19}   & 0.390 & 0.520 & 0.500 & 0.516 & 0.504 & 0.510 & 6.8\,M$^{*}$ & 11\,236\,MiB \\
 & MISA~\cite{ref17}    & 0.445 & 0.588 & 0.475 & 0.561 & 0.494 & 0.525 & \textbf{1.6\,M}$^{*}$ & \textbf{376\,MiB} \\
 & MulT~\cite{ref5}     & \underline{0.470} & 0.610 & \underline{0.511} & 0.587 & \underline{0.528} & 0.556 & \underline{4.0\,M}$^{*}$ & \underline{1\,336\,MiB} \\
 & AMP~\cite{ref23}     & \textbf{0.475} & 0.607 & 0.509 & 0.584 & 0.526 & 0.554 & 11.0\,M$^{*}$ & 11\,722\,MiB \\
 & LDDU~\cite{ref13}    & \underline{0.470} & \underline{0.624} & 0.507 & \underline{0.596} & 0.527 & \underline{0.559} & 5.97\,M & 2\,048\,MiB \\
\rowcolor{merprow}
\cellcolor{white}\multirow{-8}{*}{\shortstack{CMU-\\MOSEI}} & \textbf{MERP (ours)}  & 0.469 & \textbf{0.626} & \textbf{0.512} & \textbf{0.599} & \textbf{0.531} & \textbf{0.563} & 5.28\,M & 2\,048\,MiB \\
\midrule
 & HHMPN~\cite{ref4}    & 0.586 & 0.743 & 0.721 & 0.738 & 0.725 & 0.732 & 33.0\,M$^{*}$ & $>$15\,360\,MiB \\
 & MISA~\cite{ref17}    & 0.623 & 0.727 & 0.760 & 0.733 & 0.753 & 0.743 & \textbf{1.7\,M}$^{*}$ & \underline{1\,882\,MiB} \\
 & TAILOR~\cite{ref6}   & \underline{0.652} & 0.769 & \textbf{0.767} & 0.769 & \underline{0.767} & 0.768 & 14.7\,M$^{*}$ & 6\,604\,MiB \\
 & MulT~\cite{ref5}     & 0.651 & 0.777 & 0.762 & 0.774 & 0.765 & 0.769 & \underline{4.1\,M}$^{*}$ & \textbf{1\,408\,MiB} \\
 & AMP~\cite{ref23}     & \underline{0.652} & 0.790 & 0.761 & 0.784 & \underline{0.767} & \underline{0.775} & 11.5\,M$^{*}$ & 11\,664\,MiB \\
 & CARAT~\cite{ref19}   & 0.489 & 0.780 & 0.712 & 0.765 & 0.725 & 0.744 & 6.9\,M$^{*}$ & 11\,236\,MiB \\
 & LDDU~\cite{ref13}    & 0.614 & \textbf{0.887} & 0.662 & \textbf{0.831} & 0.697 & 0.758 & 6.41\,M & 2\,048\,MiB \\
\rowcolor{merprow}
\cellcolor{white}\multirow{-8}{*}{M\textsuperscript{3}ED} & \textbf{MERP (ours)}  & \textbf{0.659} & \underline{0.796} & \underline{0.765} & \underline{0.790} & \textbf{0.771} & \textbf{0.780} & 10.92\,M & 2\,048\,MiB \\
\bottomrule
\end{tabularx}
\end{table*}

\paragraph*{\textbf{Observations}} Three observations emerge. First, several Transformer-fusion baselines degrade sharply on CMU-MOSEI: HHMPN reaches 0.479 micro-F1 and TAILOR reaches 0.446, with TAILOR's recall falling to 0.374. Second, MERP attains the best micro-F1 on both datasets (0.563 on CMU-MOSEI and 0.780 on M\textsuperscript{3}ED), surpassing the 2024 contrastive SOTA CARAT (0.510 / 0.744) and the 2025 probabilistic SOTA LDDU (0.559 / 0.758) while using 2\,048\,MiB of inference memory. The asymmetric $F_{\beta}$ scores further show that this advantage is not tied to one precision--recall trade-off: MERP ranks first on both $F_{0.5}$ and $F_2$ on CMU-MOSEI, and first on $F_2$ and second on $F_{0.5}$ on M\textsuperscript{3}ED. Third, among the methods approaching MERP's accuracy, the Transformer-fusion baselines use approximately 3$\times$ to 7$\times$ more memory.

\paragraph*{\textbf{Clean-to-missing degradation}} Figure~\ref{fig:robustness-triptych}(a) summarises the absolute micro-F1 decrease from complete input to the $p_{\text{miss}}{=}0.2$ setting. To give the baselines their strongest clean reference, the complete-input values for CARAT~\cite{ref19} and LDDU~\cite{ref13} are taken from their published full-budget reports; their missing-input values come from our reproduction under matched preprocessing and the common missing schedule. MERP uses the single-seed checkpoint reported in Table~\ref{tab:main_results}. CARAT and LDDU lose 4.4--7.1 and 2.8--4.9 F1 points, respectively, whereas MERP loses 2.2 points on CMU-MOSEI (0.585 to 0.563) and 0.2 points on M\textsuperscript{3}ED. MERP therefore becomes the strongest model in the deployment-relevant missing regime, showing that complete-input performance alone does not predict robustness.

\subsection{Inference Cost}\label{sec:cost}
Table~\ref{tab:cost} reports per-sample inference latency and parameter counts for MERP and LDDU---the only two methods with directly comparable architectural scale---on a single A100 GPU. MERP is 27\% faster than LDDU on CMU-MOSEI and 28\% faster on M\textsuperscript{3}ED; on the smaller-feature CMU-MOSEI benchmark MERP also uses 11\% fewer parameters. On M\textsuperscript{3}ED, MERP carries more parameters because the higher-dimensional pre-trained features enlarge the input projections, while the fusion backbone remains unchanged. Both methods use 2{,}048\,MiB of inference memory, whereas the Transformer-fusion baselines in Table~\ref{tab:main_results} use approximately 3--7$\times$ more. This gap is dominated by activation cost: the latent bottleneck reduces the self-attention footprint from $O(T_{\text{total}}^2)$ to $O(LT_{\text{total}}+L^2)$ with $L\!\ll\!T_{\text{total}}$.

\begin{table}[!t]
\caption{Inference cost on a single A100 GPU. Latency is averaged over 200 single-sample forward passes after a 20-step warm-up. The best result in each column is in \textbf{bold}; lower is better, and the MERP row is highlighted in very light purple.\label{tab:cost}}
\centering
\small
\setlength{\tabcolsep}{2pt}
\renewcommand{\arraystretch}{1.08}
\begin{tabularx}{\columnwidth}{@{}
>{\hsize=1.25\hsize\raggedright\arraybackslash}X
>{\hsize=.9375\hsize\centering\arraybackslash}X
>{\hsize=.9375\hsize\centering\arraybackslash}X
>{\hsize=.9375\hsize\centering\arraybackslash}X
>{\hsize=.9375\hsize\centering\arraybackslash}X@{}}
\toprule
\textbf{Method} & \multicolumn{2}{c}{\textbf{CMU-MOSEI}} & \multicolumn{2}{c}{\textbf{M\textsuperscript{3}ED}} \\
\cmidrule(lr){2-3}\cmidrule(lr){4-5}
& \textbf{Params} & \textbf{Latency} & \textbf{Params} & \textbf{Latency} \\
\midrule
LDDU~\cite{ref13}  & 5.97\,M & 7.26\,ms & \textbf{6.41\,M} & 7.29\,ms \\
\rowcolor{merprow}
\textbf{MERP (ours)} & \textbf{5.28\,M} & \textbf{5.28\,ms} & 10.92\,M & \textbf{5.21\,ms} \\
\bottomrule
\end{tabularx}
\end{table}

\subsection{Robustness across Missing Rates on CMU-MOSEI}\label{sec:robustness}
\begin{figure*}[!t]
\centering
\includegraphics[width=\textwidth]{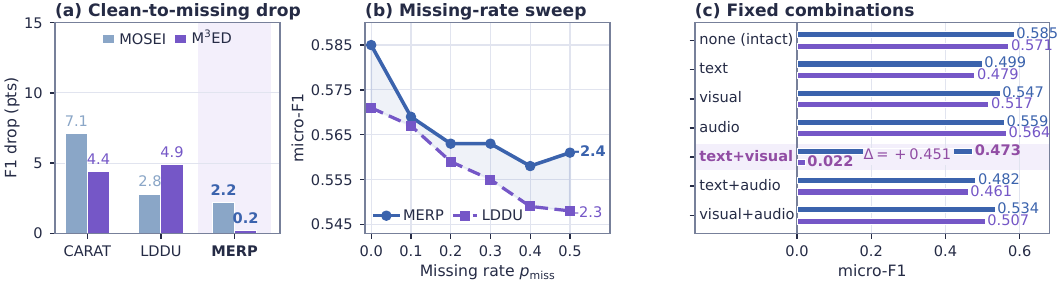}
\caption{\textbf{MERP remains robust across complementary missing-modality evaluations.} (a) Absolute micro-F1 decrease from complete input to 20\% random modality drop on CMU-MOSEI and M\textsuperscript{3}ED; lower is better. (b) Micro-F1 across increasing random-drop rates, with endpoint labels giving the absolute change from $p_{\text{miss}}{=}0$ to 0.5. (c) Exact micro-F1 under fixed missing-modality combinations applied to every CMU-MOSEI test sample. The shaded regions highlight MERP in (a) and LDDU's text--visual collapse in (c).}
\label{fig:robustness-triptych}
\end{figure*}

We sweep $p_{\text{miss}}\in\{0.0,0.1,\ldots,0.5\}$ on CMU-MOSEI, the larger of the two test sets (4{,}659 utterances), and compare MERP with LDDU under identical per-sample missing schedules. We restrict the sweep to CMU-MOSEI because M\textsuperscript{3}ED's test split contains only 179 dialogues, which is too small for a stable rate-wise comparison across six operating points. Figure~\ref{fig:robustness-triptych}(b) reports the sweep, and panel (c) summarises the fixed-combination experiment of Section~\ref{sec:per-modality}. The LDDU sweep uses one reproduced checkpoint throughout for internal consistency; its $p_{\text{miss}}{=}0$ score of 0.571 therefore differs from the published full-budget value used in panel (a).

MERP outperforms LDDU at every evaluated missing rate. The micro-F1 gap is 0.014 at $p_{\text{miss}}{=}0$, narrows to 0.002 at 0.1, and then increases to 0.013 at 0.5. MERP loses 2.4 absolute F1 points across the sweep, compared with LDDU's 2.3-point drop.

\subsection{Fixed-Combo Modality Drop}\label{sec:per-modality}
The random per-sample drop of Section~\ref{sec:robustness} mixes complete and single-missing examples. To probe more severe modality loss, we additionally evaluate every CMU-MOSEI test example with a fixed subset of modalities zeroed. We restrict this analysis to LDDU because no other published baseline reports fixed-combination results, while CARAT's clean-to-missing degradation is already shown in Figure~\ref{fig:robustness-triptych}(a). Figure~\ref{fig:robustness-triptych}(c) reports the six non-trivial combinations together with the intact reference.

\paragraph*{\textbf{LDDU collapses under text+visual drop on CMU-MOSEI}} LDDU's micro-F1 falls to 0.022 when both the text and visual streams are zeroed, while MERP retains 0.473---a 0.451 absolute gap. LDDU independently encodes each modality and merges the resulting distribution parameters at a late fusion stage; the audio-only input therefore differs substantially from its complete-input training condition. MERP applies the same input-to-latent mapping to the zeroed multi-modal tensor and performs all deeper reasoning on the fixed latent bank. With only audio content available, MERP retains 80.9\% of its complete-input F1, whereas LDDU retains less than 4\%.

\paragraph*{\textbf{Graceful degradation across all combos}} Beyond the catastrophic case, MERP outperforms LDDU on six of the seven combos (and is within 0.005 on the seventh). The single-modality drop costs MERP 2.6--8.6 absolute points; the double-modality drop costs 5.1--11.2 points but never causes a collapse, in contrast to LDDU.

\begin{figure*}[!t]
\centering
\includegraphics[width=\textwidth]{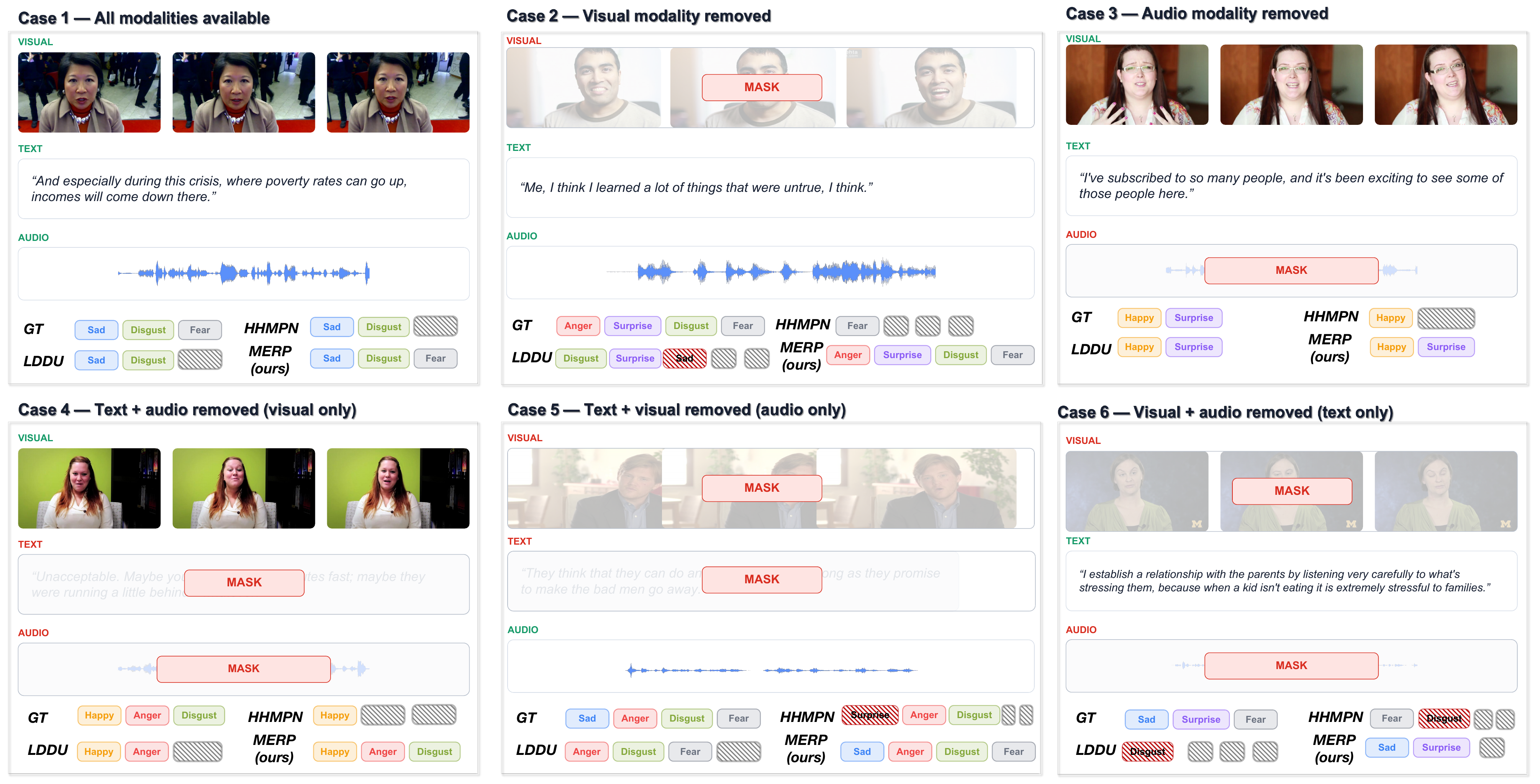}
\caption{\textbf{MERP more consistently preserves multi-label predictions across diverse modality-loss patterns.} Six CMU-MOSEI examples cover complete input, visual or audio removal, and visual-only, audio-only, or text-only views. We compare ground truth (GT) with LDDU, HHMPN, and MERP; gray hatched boxes denote missed labels, while red-hatched labels denote spurious predictions. MERP most closely matches the ground-truth label sets across the six cases, including the severe audio-only condition.}
\label{fig:case-study}
\end{figure*}

\paragraph*{\textbf{Qualitative case study}} Figure~\ref{fig:case-study} presents six representative CMU-MOSEI examples spanning complete input, single-modality removal, and single-retained views. With complete input (Case~1), MERP recovers all three emotions, whereas both baselines omit \emph{Fear}; with the visual stream removed (Case~2), it retains all four labels, while LDDU and HHMPN lose two and three, respectively. In the audio-missing example (Case~3), MERP and LDDU recover both \emph{Happy} and \emph{Surprise}, whereas HHMPN misses \emph{Surprise}. Under the more challenging visual-only and audio-only settings (Cases~4--5), MERP exactly matches the ground-truth sets, whereas the baselines omit valid labels and HHMPN additionally introduces \emph{Surprise} in the audio-only case. In the text-only example (Case~6), MERP retains \emph{Sad} and \emph{Surprise} but misses \emph{Fear}, indicating graceful rather than perfect recovery under extreme information loss. These cases suggest that MERP preserves class-specific decisions from surviving evidence instead of simply suppressing predictions, complementing the aggregate trends in Table~\ref{tab:main_results} and Figure~\ref{fig:robustness-triptych}(c).

\subsection{Ablation Study}\label{sec:ablation}

\begin{figure*}[!t]
\centering
\includegraphics[width=\textwidth]{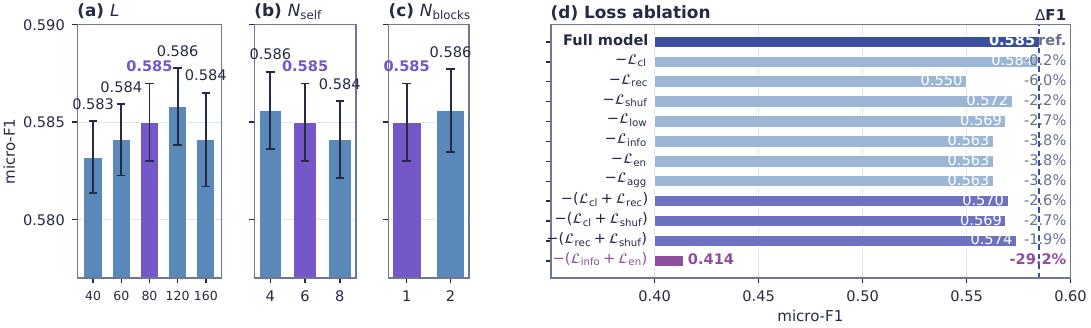}
\caption{\textbf{MERP is stable across structural settings, while reconstruction and enhancement objectives have the largest ablation effects.} (a)--(c) Structural ablations with two-seed micro-F1 means and sample-standard-deviation whiskers; purple denotes the default. (d) Loss ablations, with micro-F1 recomputed from micro-P/R. Bar labels are scores and the right column gives changes from the dashed full-model reference; plum marks the largest drop. All ablations use CMU-MOSEI with $p_{\text{miss}}{=}0$.}
\label{fig:ablation-summary}
\end{figure*}

We assess MERP's architecture and objectives on CMU-MOSEI with $p_{\text{miss}}{=}0$. Figure~\ref{fig:ablation-summary}(a)--(c) shows that varying $L$, $N_{\text{self}}$, or $N_{\text{blocks}}$ changes the two-seed mean by at most 0.31\% relative to the default (0.5850), with means ranging from 0.5832 to 0.5858 (a span of 0.0026). In panel (d), removing $\mathcal{L}_{\mathrm{rec}}$ causes the largest single-loss drop, from 0.585 to 0.550 ($-6.0\%$); removing $\mathcal{L}_{\mathrm{info}}$, $\mathcal{L}_{\mathrm{en}}$, or $\mathcal{L}_{\mathrm{agg}}$ gives 0.563 (about $-3.8\%$), while removing $\mathcal{L}_{\mathrm{cl}}$ gives 0.584 ($-0.2\%$). Removing $\mathcal{L}_{\mathrm{info}}$ and $\mathcal{L}_{\mathrm{en}}$ jointly produces the largest drop, to 0.414 ($-29.2\%$). Loss symbols follow Eqs.~\eqref{eq:total-objective} and~\eqref{eq:app-supervised-loss}.

\section{Conclusion}
This work addresses robust multi-modal multi-label emotion recognition when parts of the multi-modal evidence are unavailable. MERP compresses the cues that remain into a fixed latent workspace and preserves emotion-specific evidence through cross-level readout and reconstruction, prototype-guided noise-gated enhancement, and confidence-aware class-wise fusion. Experiments across two benchmarks and complementary modality-drop settings show that this design achieves leading recognition performance while degrading gracefully under increasingly severe and compound modality loss. These results suggest a broader lesson for robust multi-modal learning: protecting task-relevant evidence after compact aggregation can be more effective than reconstructing missing streams or assigning a single reliability estimate to each modality. Building on this principle, future work can investigate adaptive latent workspaces that retain bounded deep reasoning while supporting evolving label spaces and longer multi-modal contexts.

\bibliographystyle{IEEEtran}
\bibliography{ref}

\begin{thebibliography}{10}
\providecommand{\url}[1]{#1}
\csname url@samestyle\endcsname
\providecommand{\newblock}{\relax}
\providecommand{\bibinfo}[2]{#2}
\providecommand{\BIBentrySTDinterwordspacing}{\spaceskip=0pt\relax}
\providecommand{\BIBentryALTinterwordstretchfactor}{4}
\providecommand{\BIBentryALTinterwordspacing}{\spaceskip=\fontdimen2\font plus
\BIBentryALTinterwordstretchfactor\fontdimen3\font minus
  \fontdimen4\font\relax}
\providecommand{\BIBforeignlanguage}[2]{{%
\expandafter\ifx\csname l@#1\endcsname\relax
\typeout{** WARNING: IEEEtran.bst: No hyphenation pattern has been}%
\typeout{** loaded for the language `#1'. Using the pattern for}%
\typeout{** the default language instead.}%
\else
\language=\csname l@#1\endcsname
\fi
#2}}
\providecommand{\BIBdecl}{\relax}
\BIBdecl

\bibitem{ref1}
S.~Poria, E.~Cambria, R.~Bajpai, and A.~Hussain, ``A review of affective
  computing: From unimodal analysis to multimodal fusion,'' \emph{Information
  Fusion}, vol.~37, pp. 98--125, 2017.

\bibitem{ref2}
P.~K. Atrey, M.~A. Hossain, A.~E. Saddik, and M.~S. Kankanhalli, ``Multimodal
  fusion for multimedia analysis: A survey,'' \emph{Multimedia Systems},
  vol.~16, no.~6, pp. 345--379, 2010.

\bibitem{ref3}
D.~Zhang, X.~Ju, J.~Li, S.~Li, Q.~Zhu, and G.~Zhou, ``Multi-modal multi-label
  emotion detection with modality and label dependence,'' in \emph{Proceedings
  of the 2020 Conference on Empirical Methods in Natural Language Processing
  (EMNLP)}.\hskip 1em plus 0.5em minus 0.4em\relax Association for
  Computational Linguistics, 2020, pp. 3584--3593.

\bibitem{ref4}
D.~Zhang, X.~Ju, W.~Zhang, J.~Li, S.~Li, Q.~Zhu, and G.~Zhou, ``Multi-modal
  multi-label emotion recognition with heterogeneous hierarchical message
  passing,'' in \emph{Proceedings of the AAAI Conference on Artificial
  Intelligence}, vol.~35, no.~16, 2021, pp. 14\,338--14\,346.

\bibitem{ref6}
Y.~Zhang, M.~Chen, J.~Shen, and C.~Wang, ``{TAILOR}: Versatile multi-modal
  learning for multi-label emotion recognition,'' in \emph{Proceedings of the
  AAAI Conference on Artificial Intelligence}, vol.~36, no.~8, 2022, pp.
  9100--9108.

\bibitem{ref19}
C.~Peng, K.~Chen, L.~Shou, and G.~Chen, ``{CARAT}: Contrastive feature
  reconstruction and aggregation for multi-modal multi-label emotion
  recognition,'' in \emph{Proceedings of the AAAI Conference on Artificial
  Intelligence}, vol.~38, no.~13, 2024, pp. 14\,581--14\,589.

\bibitem{ref13}
J.~Huang, J.~Zhong, Q.~Lei, J.~Gao, Y.~Yang, S.~Wang, P.~Li, and K.~Wei,
  ``Latent distribution decouple for uncertain-aware multimodal multi-label
  emotion recognition,'' in \emph{Findings of the Association for Computational
  Linguistics: ACL 2025}.\hskip 1em plus 0.5em minus 0.4em\relax Association
  for Computational Linguistics, 2025, pp. 24\,123--24\,138.

\bibitem{ref35}
J.~Li, X.~Wang, Y.~Liu, and Z.~Zeng, ``{CFN-ESA}: A cross-modal fusion network
  with emotion-shift awareness for dialogue emotion recognition,'' \emph{IEEE
  Transactions on Affective Computing}, vol.~15, no.~4, pp. 1919--1933, 2024.

\bibitem{ref36}
Y.~Wang, W.~Zhang, K.~Liu, W.~Wu, F.~Hu, H.~Yu, and G.~Wang, ``Dynamic
  emotion-dependent network with relational subgraph interaction for multimodal
  emotion recognition,'' \emph{IEEE Transactions on Affective Computing},
  vol.~16, no.~2, pp. 712--725, 2025.

\bibitem{ref37}
P.~Kumar, S.~Malik, B.~Raman, and X.~Li, ``{VISTANet}: {VIsual Spoken Textual
  Additive Net} for interpretable multimodal emotion recognition,'' \emph{IEEE
  Transactions on Affective Computing}, vol.~16, no.~4, pp. 2881--2893, 2025.

\bibitem{ref5}
Y.-H.~H. Tsai, S.~Bai, P.~P. Liang, J.~Z. Kolter, L.-P. Morency, and
  R.~Salakhutdinov, ``Multimodal transformer for unaligned multimodal language
  sequences,'' in \emph{Proceedings of the 57th Annual Meeting of the
  Association for Computational Linguistics}.\hskip 1em plus 0.5em minus
  0.4em\relax Association for Computational Linguistics, 2019, pp. 6558--6569.

\bibitem{ref21}
J.~Zhao, R.~Li, and Q.~Jin, ``Missing modality imagination network for emotion
  recognition with uncertain missing modalities,'' in \emph{Proceedings of the
  59th Annual Meeting of the Association for Computational Linguistics and the
  11th International Joint Conference on Natural Language Processing (Volume 1:
  Long Papers)}.\hskip 1em plus 0.5em minus 0.4em\relax Association for
  Computational Linguistics, 2021, pp. 2608--2618.

\bibitem{ref24}
R.~Lin and H.~Hu, ``{MissModal}: Increasing robustness to missing modality in
  multimodal sentiment analysis,'' \emph{Transactions of the Association for
  Computational Linguistics}, vol.~11, pp. 1686--1702, 2023.

\bibitem{ref30}
A.~Zhu, M.~Hu, X.~Wang, J.~Yang, Y.~Tang, and N.~An, ``Proxy-driven robust
  multimodal sentiment analysis with incomplete data,'' in \emph{Proceedings of
  the 63rd Annual Meeting of the Association for Computational Linguistics
  (Volume 1: Long Papers)}.\hskip 1em plus 0.5em minus 0.4em\relax Association
  for Computational Linguistics, 2025, pp. 22\,123--22\,138.

\bibitem{ref31}
M.~K. Reza, A.~Prater-Bennette, and M.~S. Asif, ``Robust multimodal learning
  with missing modalities via parameter-efficient adaptation,'' \emph{IEEE
  Transactions on Pattern Analysis and Machine Intelligence}, vol.~47, no.~2,
  pp. 742--754, 2025.

\bibitem{ref38}
K.~Zeng, F.~Chen, J.~Shao, and H.~T. Shen, ``Heterogeneous graph denoising
  network for incomplete multimodal emotion recognition in conversation,''
  \emph{IEEE Transactions on Affective Computing}, pp. 1--14, 2026, early
  access.

\bibitem{ref34}
\BIBentryALTinterwordspacing
R.~Wu, H.~Wang, H.-T. Chen, and G.~Carneiro, ``Deep multimodal learning with
  missing modality: A survey,'' \emph{Transactions on Machine Learning
  Research}, 2026. [Online]. Available:
  \url{https://openreview.net/forum?id=tc7RFcx4hT}
\BIBentrySTDinterwordspacing

\bibitem{ref9}
\BIBentryALTinterwordspacing
A.~Jaegle, S.~Borgeaud, J.-B. Alayrac, C.~Doersch, C.~Ionescu, D.~Ding,
  S.~Koppula, D.~Zoran, A.~Brock, E.~Shelhamer, O.~J. H{\'e}naff, M.~M.
  Botvinick, A.~Zisserman, O.~Vinyals, and J.~Carreira, ``{Perceiver IO}: A
  general architecture for structured inputs and outputs,'' in
  \emph{International Conference on Learning Representations}, 2022. [Online].
  Available: \url{https://openreview.net/forum?id=fILj7WpI-g}
\BIBentrySTDinterwordspacing

\bibitem{ref18}
X.~Ju, D.~Zhang, J.~Li, and G.~Zhou, ``Transformer-based label set generation
  for multi-modal multi-label emotion detection,'' in \emph{Proceedings of the
  28th ACM International Conference on Multimedia}.\hskip 1em plus 0.5em minus
  0.4em\relax Association for Computing Machinery, 2020, pp. 512--520.

\bibitem{ref23}
S.~Ge, Z.~Jiang, Z.~Cheng, C.~Wang, Y.~Yin, and Q.~Gu, ``Learning robust
  multi-modal representation for multi-label emotion recognition via
  adversarial masking and perturbation,'' in \emph{Proceedings of the ACM Web
  Conference 2023}.\hskip 1em plus 0.5em minus 0.4em\relax Association for
  Computing Machinery, 2023, pp. 1510--1518.

\bibitem{ref22}
W.~Zheng, J.~Yu, and R.~Xia, ``A unimodal valence-arousal driven contrastive
  learning framework for multimodal multi-label emotion recognition,'' in
  \emph{Proceedings of the 32nd ACM International Conference on
  Multimedia}.\hskip 1em plus 0.5em minus 0.4em\relax Association for Computing
  Machinery, 2024, pp. 622--631.

\bibitem{ref28}
D.~Hu, X.~Jiang, Z.~Sun, F.~Shen, and X.~Xu, ``Heterogeneous graph embedding
  for multimodal multi-label emotion recognition,'' in \emph{Proceedings of the
  2025 International Conference on Multimedia Retrieval}.\hskip 1em plus 0.5em
  minus 0.4em\relax Association for Computing Machinery, 2025, pp. 460--468.

\bibitem{ref20}
M.~Ma, J.~Ren, L.~Zhao, S.~Tulyakov, C.~Wu, and X.~Peng, ``{SMIL}: Multimodal
  learning with severely missing modality,'' in \emph{Proceedings of the AAAI
  Conference on Artificial Intelligence}, vol.~35, no.~3, 2021, pp. 2302--2310.

\bibitem{ref26}
Z.~Yuan, W.~Li, H.~Xu, and W.~Yu, ``Transformer-based feature reconstruction
  network for robust multimodal sentiment analysis,'' in \emph{Proceedings of
  the 29th ACM International Conference on Multimedia}.\hskip 1em plus 0.5em
  minus 0.4em\relax Association for Computing Machinery, 2021, pp. 4400--4407.

\bibitem{ref41}
R.~Dai, C.~Li, Y.~Yan, L.~Mo, K.~Qin, and T.~He, ``Unbiased missing-modality
  multimodal learning,'' in \emph{Proceedings of the IEEE/CVF International
  Conference on Computer Vision}, 2025, pp. 24\,507--24\,517.

\bibitem{ref42}
G.~Ke, S.~He, X.~Wang, B.~Wang, G.~Chao, Y.~Zhang, Y.~Xie, and H.~Su,
  ``Knowledge bridger: Towards training-free missing modality completion,'' in
  \emph{Proceedings of the IEEE/CVF Conference on Computer Vision and Pattern
  Recognition}, 2025, pp. 25\,864--25\,873.

\bibitem{ref32}
G.~Chochlakis, T.~Iqbal, W.~H. Kang, and Z.~Huang, ``Modality-agnostic
  multimodal emotion recognition using a contrastive masked autoencoder,'' in
  \emph{Proceedings of Interspeech 2025}, 2025, pp. 3005--3009.

\bibitem{ref40}
S.~Li, C.~Chen, and J.~Han, ``{SimMLM}: A simple framework for multi-modal
  learning with missing modality,'' in \emph{Proceedings of the IEEE/CVF
  International Conference on Computer Vision}, 2025, pp. 24\,068--24\,077.

\bibitem{ref17}
D.~Hazarika, R.~Zimmermann, and S.~Poria, ``{MISA}: Modality-invariant and
  -specific representations for multimodal sentiment analysis,'' in
  \emph{Proceedings of the 28th ACM International Conference on
  Multimedia}.\hskip 1em plus 0.5em minus 0.4em\relax Association for Computing
  Machinery, 2020, pp. 1122--1131.

\bibitem{ref27}
\BIBentryALTinterwordspacing
A.~Nagrani, S.~Yang, A.~Arnab, A.~Jansen, C.~Schmid, and C.~Sun, ``Attention
  bottlenecks for multimodal fusion,'' in \emph{Advances in Neural Information
  Processing Systems}, vol.~34, 2021, pp. 14\,200--14\,213. [Online].
  Available:
  \url{https://proceedings.neurips.cc/paper/2021/hash/76ba9f564ebbc35b1014ac498fafadd0-Abstract.html}
\BIBentrySTDinterwordspacing

\bibitem{ref29}
Z.~Gao, X.~Jiang, X.~Xu, F.~Shen, Y.~Li, and H.~T. Shen, ``Embracing unimodal
  aleatoric uncertainty for robust multimodal fusion,'' in \emph{Proceedings of
  the IEEE/CVF Conference on Computer Vision and Pattern Recognition}, 2024,
  pp. 26\,866--26\,875.

\bibitem{ref33}
R.~Lin, Q.~He, S.~Mai, Y.~Zeng, A.~Xiong, L.~Huang, Y.-P. Tan, and H.~Hu,
  ``{CyIN}: Cyclic informative latent space for bridging complete and
  incomplete multimodal learning,'' in \emph{Advances in Neural Information
  Processing Systems}, vol.~38, 2025.

\bibitem{ref39}
\BIBentryALTinterwordspacing
S.~Kornblith, M.~Norouzi, H.~Lee, and G.~Hinton, ``Similarity of neural network
  representations revisited,'' in \emph{Proceedings of the 36th International
  Conference on Machine Learning}, ser. Proceedings of Machine Learning
  Research, vol.~97.\hskip 1em plus 0.5em minus 0.4em\relax PMLR, 2019, pp.
  3519--3529. [Online]. Available:
  \url{https://proceedings.mlr.press/v97/kornblith19a.html}
\BIBentrySTDinterwordspacing

\bibitem{ref10}
P.~Khosla, P.~Teterwak, C.~Wang, A.~Sarna, Y.~Tian, P.~Isola, A.~Maschinot,
  C.~Liu, and D.~Krishnan, ``Supervised contrastive learning,'' in
  \emph{Advances in Neural Information Processing Systems}, vol.~33, 2020, pp.
  18\,661--18\,673.

\bibitem{ref11}
A.~B. Zadeh, P.~P. Liang, S.~Poria, E.~Cambria, and L.-P. Morency, ``Multimodal
  language analysis in the wild: {CMU-MOSEI} dataset and interpretable dynamic
  fusion graph,'' in \emph{Proceedings of the 56th Annual Meeting of the
  Association for Computational Linguistics (Volume 1: Long Papers)}.\hskip 1em
  plus 0.5em minus 0.4em\relax Association for Computational Linguistics, 2018,
  pp. 2236--2246.

\bibitem{ref12}
J.~Zhao, T.~Zhang, J.~Hu, Y.~Liu, Q.~Jin, X.~Wang, and H.~Li, ``{M3ED}:
  Multi-modal multi-scene multi-label emotional dialogue database,'' in
  \emph{Proceedings of the 60th Annual Meeting of the Association for
  Computational Linguistics (Volume 1: Long Papers)}.\hskip 1em plus 0.5em
  minus 0.4em\relax Association for Computational Linguistics, 2022, pp.
  5699--5710.

\bibitem{ref25}
J.~R. Pierce, \emph{An Introduction to Information Theory: Symbols, Signals and
  Noise}.\hskip 1em plus 0.5em minus 0.4em\relax New York, NY, USA: Dover
  Publications, 1980.

\end{thebibliography}

\clearpage
\appendices
\section{Preprocessing and Attention Details}\label{app:preprocessing}

\subsection{Fourier Positions and Channel Alignment}
For a sequence of length $T_m$, the scalar position $x_i$ is linearly spaced on $[-1,1]$. With $K{=}32$ frequency bands $f_k$ linearly spaced between $1$ and $T_m/2$, the position vector is
\begin{equation}
\mathbf{p}_i=\left[x_i,\{\sin(\pi f_kx_i)\}_{k=1}^{K},
\{\cos(\pi f_kx_i)\}_{k=1}^{K}\right]\in\mathbb{R}^{2K+1}.
\label{eq:app-fourier}
\end{equation}
The content-position representation is
\begin{equation}
\mathbf{E}_{m}=[\operatorname{LN}(\bar{\mathbf{X}}_m);\mathbf{P}_m]
\in\mathbb{R}^{B\times T_m\times(d_m+2K+1)}.
\label{eq:app-content-position}
\end{equation}
Because the input dimensions differ across modalities, trainable padding embeddings extend each $\mathbf{E}_m$ to the common width
$D=\max_{m\in\mathcal{M}}(d_m+2K+1)$. The modality-specific linear gate in Eq.~\eqref{eq:modality-gate} is applied after this channel alignment. Missingness zeros the content term in Eq.~\eqref{eq:app-content-position}; the deterministic position code is retained.

\subsection{Attention Blocks}\label{app:attention}
For one attention head with query width $d_h$, scaled dot-product attention is
\begin{equation}
\operatorname{Attn}(\mathbf{Q},\mathbf{K},\mathbf{V})=
\operatorname{softmax}\!\left(
\frac{(\mathbf{Q}\mathbf{W}_Q)(\mathbf{K}\mathbf{W}_K)^{\top}}
{\sqrt{d_h}}
\right)(\mathbf{V}\mathbf{W}_V).
\label{eq:app-attention}
\end{equation}
Multi-head outputs are concatenated and linearly projected. Each cross-attention and self-attention block uses pre-layer normalisation, a residual attention update, and a residual feed-forward update:
\begin{equation}
\begin{aligned}
\mathbf{H}'&=\mathbf{H}+\operatorname{MHA}(\operatorname{LN}(\mathbf{H}),\operatorname{LN}(\mathbf{S}),\operatorname{LN}(\mathbf{S})),\\
\operatorname{Block}(\mathbf{H},\mathbf{S})&=\mathbf{H}'+
\operatorname{MLP}(\operatorname{LN}(\mathbf{H}')).
\end{aligned}
\label{eq:app-attention-block}
\end{equation}
For encoder cross-attention, $\mathbf{H}=\mathbf{Z}^{(0)}$ and $\mathbf{S}=\mathbf{U}$. Self-attention sets $\mathbf{S}=\mathbf{H}$. The encoder uses one cross-attention head and eight self-attention heads. The decoder applies Eq.~\eqref{eq:app-attention-block} with $C$ broadcast trainable queries as $\mathbf{H}$ and $\mathbf{Z}_{\mathrm{high}}$ as $\mathbf{S}$, followed by a linear projection to width $H$.

\section{Prototype and Contrastive Learning Details}\label{app:prototypes}

\subsection{Dynamic Class Prototypes}
The model maintains $2C$ prototype buffers $\{\boldsymbol{\mu}_{c}^{+},\boldsymbol{\mu}_{c}^{-}\}_{c=1}^{C}$ in the $d_z$-dimensional projection space. For a training pair $(b,c)$, the selected sign is $s_{b,c}=+$ when $Y_{b,c}=1$ and $s_{b,c}=-$ when $Y_{b,c}=0$. Its retention coefficient is
\begin{equation}
m_{b,c}=m_0\frac{1+cos(\mathbf{h}_{b,c},
\boldsymbol{\mu}_{c}^{s_{b,c}})}{2},\qquad m_0=0.99,
\label{eq:app-proto-momentum}
\end{equation}
and the selected prototype is updated sequentially within the mini-batch:
\begin{equation}
\boldsymbol{\mu}_{c}^{s_{b,c}}\leftarrow
m_{b,c}\boldsymbol{\mu}_{c}^{s_{b,c}}+(1-m_{b,c})\mathbf{h}_{b,c}.
\label{eq:app-proto-update}
\end{equation}
The opposite-sign prototype is unchanged for that pair. Equation~\eqref{eq:prototype-mixture} uses both class prototypes during training. At inference, the sign with the larger prototype score is selected before the inverse projection $\psi$.

\subsection{Queue-Based Supervised Contrastive Loss}
Two class-conditioned views are used for contrastive learning: the projected high-level feature $\phi(\mathbf{F}_{\mathrm{high}})$ and the projected enhanced feature $\phi(\mathbf{F}_{\mathrm{en}})$. After $\ell_2$ normalisation, each view receives the pseudo-label
\begin{equation}
\ell_{b,c}=
\begin{cases}
c, & Y_{b,c}=1,\\
C+c, & Y_{b,c}=0.
\end{cases}
\label{eq:app-pseudo-label}
\end{equation}
A first-in--first-out queue of 768 feature--label pairs supplies additional comparisons. Let $\mathcal{A}$ denote current anchors together with queued features, $\mathcal{P}(i)=\{p\neq i:\ell_p=\ell_i\}$, and $\tau_{\mathrm{cl}}=0.07$. The supervised contrastive loss is
\begin{equation}
\mathcal{L}_{\mathrm{cl}}=-\frac{1}{|\mathcal{A}|}
\sum_{i\in\mathcal{A}}\frac{1}{|\mathcal{P}(i)|}
\sum_{p\in\mathcal{P}(i)}
\log\frac{\exp(\mathbf{h}_i^{\top}\mathbf{h}_p/\tau_{\mathrm{cl}})}
{\sum_{a\in\mathcal{A}\setminus\{i\}}
\exp(\mathbf{h}_i^{\top}\mathbf{h}_a/\tau_{\mathrm{cl}})}.
\label{eq:app-supcon}
\end{equation}
Anchors without a positive partner contribute zero to the batch mean.

\section{Spectral Noise-Gated Feature Block}\label{app:snga}
Let $D_e=3H/2$. The concatenated feature in Eq.~\eqref{eq:enhance-input} is first reduced:
\begin{equation}
\mathbf{R}=\operatorname{Dropout}\!\left(
\operatorname{ReLU}(\operatorname{LN}(\mathbf{W}_r\mathbf{F}_{\mathrm{cat}}+\mathbf{b}_r))
\right)\in\mathbb{R}^{B\times C\times D_e}.
\label{eq:app-snga-reduce}
\end{equation}
The feature response and gate are
\begin{equation}
\mathbf{A}=\tanh(\mathbf{W}_a\mathbf{R}+\mathbf{b}_a)+\boldsymbol{\epsilon},
\label{eq:app-snga-response}
\end{equation}
\begin{equation}
\mathbf{G}=\mathbf{R}\odot\sigma\!\left(
\mathbf{W}_{g2}\operatorname{ReLU}(\mathbf{W}_{g1}\mathbf{R}+\mathbf{b}_{g1})+\mathbf{b}_{g2}
\right).
\label{eq:app-snga-gate}
\end{equation}
The output is
\begin{equation}
\mathbf{F}_{\mathrm{en}}=
\operatorname{LN}\!\left(
\mathbf{W}_o\operatorname{LN}
(\operatorname{Dropout}(\mathbf{A}\odot\mathbf{G})+\mathbf{R})+\mathbf{b}_o
\right).
\label{eq:app-snga-output}
\end{equation}

During training, $\boldsymbol{\epsilon}$ is pink feature noise generated independently for each sample and class~\cite{ref25}. For white noise $\boldsymbol{\xi}$, let $\widehat{\boldsymbol{\xi}}=\operatorname{RFFT}(\boldsymbol{\xi})$ along the last feature axis and define $q_0=1$, $q_k=k^{-1/2}$ for $k\geq1$. The injected noise is
\begin{equation}
\boldsymbol{\epsilon}=\rho\,
\frac{\operatorname{IRFFT}(\mathbf{q}\odot\widehat{\boldsymbol{\xi}})}
{\operatorname{Std}[\operatorname{IRFFT}(\mathbf{q}\odot\widehat{\boldsymbol{\xi}})]},\qquad
\rho=0.1.
\label{eq:app-pink-noise}
\end{equation}
This filter produces a $1/f$ power spectrum over feature channels. At inference, $\boldsymbol{\epsilon}=\mathbf{0}$. The block regulariser is the sum of Frobenius norms of $\mathbf{W}_a$, $\mathbf{W}_{g1}$, $\mathbf{W}_{g2}$, and $\mathbf{W}_o$.

\section{Losses and Fusion Details}\label{app:objective}

\subsection{Classification Losses}
For logits $\mathbf{z}$ and targets $\mathbf{Y}$, define the mean binary cross-entropy
\begin{equation}
\begin{aligned}
\mathcal{B}(\mathbf{z},\mathbf{Y})
=-\frac{1}{BC}\sum_{b,c}\big[&Y_{b,c}\log\sigma(z_{b,c})\\
&+(1-Y_{b,c})\log(1-\sigma(z_{b,c}))\big].
\end{aligned}
\label{eq:app-bce-logit}
\end{equation}
The supervised branch loss is
\begin{equation}
\begin{aligned}
\mathcal{L}_{\mathrm{sup}}={}&
\lambda_{\mathrm{low}}\mathcal{B}(\mathbf{z}^{\mathrm{low}},\mathbf{Y})
+\lambda_{\mathrm{high}}\mathcal{B}(\mathbf{z}^{\mathrm{high}},\mathbf{Y})\\
&+\lambda_{\mathrm{en}}\mathcal{B}(\mathbf{z}^{\mathrm{en}},\mathbf{Y})
+\lambda_{\mathrm{agg}}\mathcal{B}(\mathbf{z}^{\mathrm{agg}},\mathbf{Y}).
\end{aligned}
\label{eq:app-supervised-loss}
\end{equation}
The fused-output loss is binary cross-entropy on Eq.~\eqref{eq:score-fusion}:
\begin{equation}
\begin{aligned}
\mathcal{L}_{\mathrm{final}}=-\frac{1}{BC}\sum_{b,c}\big[
&Y_{b,c}\log p^{\mathrm{final}}_{b,c}\\
&+(1-Y_{b,c})\log(1-p^{\mathrm{final}}_{b,c})\big].
\end{aligned}
\label{eq:app-final-loss}
\end{equation}
The gate $G_{\theta}$ is a $3\!\rightarrow\!32\!\rightarrow\!3$ multilayer perceptron with a ReLU hidden activation; its gradients are supplied by $\mathcal{L}_{\mathrm{final}}$.

\subsection{Feature-Consistency and Shuffle Losses}
The feature-consistency loss aligns the low-level and enhanced representations. After flattening the class and channel dimensions and applying $\ell_2$ normalisation, let $\mathbf{u}_b=\operatorname{norm}(\operatorname{vec}(\mathbf{F}_{\mathrm{en},b}))$ and $\mathbf{v}_b=\operatorname{norm}(\operatorname{vec}(\mathbf{F}_{\mathrm{low},b}))$. With $\tau_{\mathrm{info}}=0.07$,
\begin{equation}
\mathcal{L}_{\mathrm{info}}=-\frac{1}{B}\sum_{b=1}^{B}
\log\frac{\exp(\mathbf{u}_b^{\top}\mathbf{v}_b/\tau_{\mathrm{info}})}
{\sum_{k=1}^{B}\exp(\mathbf{u}_b^{\top}\mathbf{v}_k/\tau_{\mathrm{info}})}.
\label{eq:app-info}
\end{equation}

For shuffle regularisation, an independent permutation $\pi_c$ is sampled across the batch for every class. Define $\tilde{\mathbf{F}}_{b,c}=\mathbf{F}_{\mathrm{en},\pi_c(b),c}$ and $\tilde{Y}_{b,c}=Y_{\pi_c(b),c}$. Applying the aggregate classifier $h_{\mathrm{agg}}$ to the shuffled class features gives
\begin{equation}
\mathcal{L}_{\mathrm{shuf}}=
\mathcal{B}\!\left(
h_{\mathrm{agg}}(\operatorname{vec}(\tilde{\mathbf{F}})),
\tilde{\mathbf{Y}}
\right).
\label{eq:app-shuffle}
\end{equation}

\subsection{Objective Coefficients}
The coefficients used in Eq.~\eqref{eq:total-objective} are
\begin{equation}
\begin{gathered}
\lambda_{\mathrm{low}}=0.01,\quad
\lambda_{\mathrm{high}}=\lambda_{\mathrm{en}}=0.1,\quad
\lambda_{\mathrm{agg}}=1,\\
\lambda_{\mathrm{rec}}=\lambda_{\mathrm{cl}}=
\lambda_{\mathrm{info}}=1,\quad
\lambda_{\mathrm{shuf}}=0.1,\quad
\lambda_{\mathrm{reg}}=10^{-4}.
\end{gathered}
\label{eq:app-loss-weights}
\end{equation}
The coefficient of $\mathcal{L}_{\mathrm{final}}$ is one.

\end{document}